\documentclass[iicol,sn-mathphys-ay]{sn-jnl}

\usepackage{graphicx}%
\usepackage{multirow}%
\usepackage{amsmath,amssymb,amsfonts}%
\usepackage{amsthm}%
\usepackage{mathrsfs}%
\usepackage[title]{appendix}%
\usepackage{xcolor}%
\usepackage{textcomp}%
\usepackage{manyfoot}%
\usepackage{booktabs}%
\usepackage{algorithm}%
\usepackage{algorithmicx}%
\usepackage{algpseudocode}%
\usepackage{listings}%
\usepackage{xcolor}
\usepackage{array}
\usepackage{diagbox}
\usepackage[]{hyperref}
\usepackage[protrusion=true,expansion=true]{microtype}
\usepackage[T1]{fontenc}
\usepackage{lmodern}
\hypersetup{
  pdftitle={Vibro-Sense: Robust Vibration-based Impulse Response Localization and Trajectory Tracking for Robotic Hands},
  pdfsubject={Robotics (cs.RO); Artificial Intelligence (cs.AI);Force and Tactile Sensing; Perception for Grasping and Manipulation; Hardware-Software Integration in Robotics},%
  pdfkeywords={Tactile sensors},%
  colorlinks=true,
  linktoc=all,
  linkcolor=darkgray,
  citecolor=darkgray,
  urlcolor=black,
}

\theoremstyle{thmstyleone}%
\theoremstyle{thmstyletwo}%

\theoremstyle{thmstylethree}%

\begin{document}

\title[Article Title]{Vibro-Sense: Robust Vibration-based Impulse Response Localization and Trajectory Tracking for Robotic Hands}

\author*[1]{\fnm{Wadhah} \sur{Zai El Amri}}\email{wadhah.zai@l3s.de}

\author[1]{\fnm{Nicol\'as} \sur{Navarro-Guerrero}}\email{nicolas.navarro.guerrero@gmail.com}

\affil[1]{\orgdiv{Leibniz Universität Hannover}, \orgname{L3S Research Center}, \orgaddress{\street{Appelstra\ss e 4}, \city{Hanover}, \postcode{30167}, \country{Germany}}}

\abstract{
Rich contact perception is crucial for robotic manipulation, yet traditional tactile skins remain expensive and complex to integrate. This paper presents a scalable alternative: high-accuracy whole-hand touch localization via vibro-acoustic sensing. By equipping a robotic hand with seven low-cost piezoelectric microphones and leveraging an Audio Spectrogram Transformer, we decode the vibrational signatures generated during physical interaction. Extensive evaluation across stationary and dynamic tasks reveals an average localization error below 5 mm in static conditions. Furthermore, our analysis highlights the distinct influence of material properties: stiff materials (e.g., metal) excel in impulse response localization due to sharp, high-bandwidth responses, whereas textured materials (e.g., wood) provide superior friction-based features for trajectory tracking. The system demonstrates robustness to the robot's own motion, maintaining effective tracking even during active operation. Trained with additional no-contact recordings of the moving hand, the model further distinguishes external contacts from internal motor and structural vibrations, reaching a detection accuracy of 99.0\% with a false-positive rate of 0.2\%. Our focus is on contact perception, enabling a robot to sense interactions with its surroundings. Our primary contribution is demonstrating that complex physical contact dynamics can be effectively decoded from simple vibrational signals, offering a viable pathway to widespread, affordable contact perception in robotics. To accelerate research, we provide our full datasets, models, and experimental setups as open-source resources.}


\maketitle

\section{Introduction}

Humans possess a comprehensive system for perceiving their physical surroundings that extends beyond exteroceptive senses like vision and hearing. The somatosensory system provides crucial information about direct contact and environmental forces~\citep{Abraira2013Sensory,Navarro-Guerrero2023VisuoHaptic}. When an object contacts with the body, the resulting mechanical vibrations propagate through the skin and tissues, enabling immediate localization and characterization of the interaction. This distributed vibro-acoustic sensing is fundamental not only for object manipulation but for overall spatial awareness and safe interaction within a dynamic environment. The ability to detect and interpret these contact events is a key component of effective physical interaction~\citep{Wang2021Human}.

Drawing from this biological paradigm, the field of robot perception is increasingly exploring structure-borne sound analysis to supplement conventional vision and force modalities \citep{JuinaQuilachamin2023Fingerprint, Bonner2021AU, Toprak2018Evaluating}. The integration of piezoelectric sensors, particularly contact microphones, presents a promising avenue for capturing the complex dynamics of robot-environment interactions. By leveraging the unique properties of acoustic signals, researchers have made notable progress in developing vibro-acoustic sensing techniques for contact-rich manipulation~\citep{Lu2023Active,Mejia2024Hearing,Liu2024SonicSensea,Wall2022Virtual}.

In this paper, we propose a cost-effective yet accurate method that enables robots to perceive physical contact in a more natural manner. We demonstrate our approach on two real-world tasks: impulse response localization and trajectory tracking. Our method utilizes a robotic hand equipped with seven contact microphones to capture vibrational signals, which are then processed using an Audio Spectrogram Transformer (AST) architecture to predict the positions of external touches on the hand. To train and evaluate this system, we collected extensive datasets consisting of over 65,000 unique samples for impulse response localization and over 240,000 interactions for trajectory tracking.

Our contribution is two-fold. Firstly, we present a robust method for vibro-acoustic sensing that can be applied to complex robotic geometries. We provide a detailed analysis of how material properties, specifically stiffness versus texture, distinctly influence localization accuracy across different interaction modes. Secondly, we demonstrate the effectiveness of our approach in dynamic scenarios, where the robot hand actively moves while interacting with its surroundings. Our results show that the system maintains effective localization accuracy even in the presence of significant motion and actuator noise. By leveraging vibro-acoustic sensing and deep learning, our work provides a scalable solution for whole-hand contact perception. To facilitate further research, we open-source all our code, datasets, and experimental setups on our website, allowing researchers and practitioners to easily replicate and build upon our work:~\href{https://wzaielamri.github.io/publication/vibrosense}{\textcolor{gray}{wzaielamri.github.io/publication/vibrosense}}. The website also hosts demonstration videos of the real-time pipeline: impulse-response localization while the hand is powered on and its fan and motors are running, and trajectory tracking with contact detection while the hand is in active motion.

\begin{figure}[htbp]
  \centering
  \includegraphics[height=65mm]{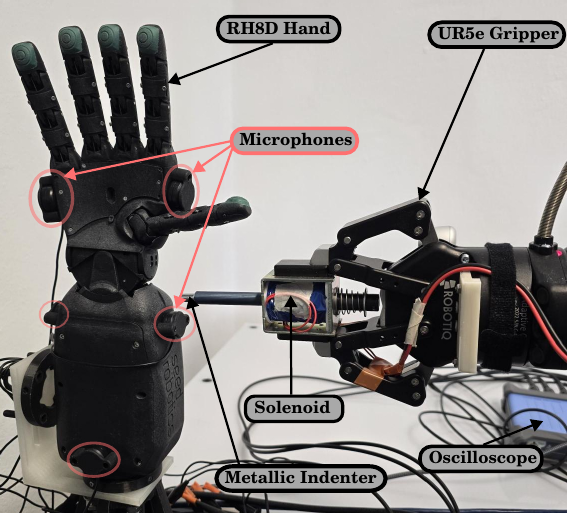}
  \caption{Setup of the impulse response localization task. Key components are labeled, including the RH8D hand and its surface-mounted microphones (circled in red). The visible microphones are identified as IDs 1 through 5, ordered from right to left and top to bottom. The UR5e gripper holds a solenoid actuator equipped with a metallic indenter to apply controlled mechanical impulses to the hand. The oscilloscope used for the experimental setup is also indicated.}
  \label{fig:experiment_setup}
\end{figure}

\section{Related Works}

For robots to operate safely and effectively in the physical world, they must be able to perceive and interpret contact within their environment. This sense of touch is crucial for enabling robots to securely manipulate objects and avoid unexpected or damaging collisions \citep{Hoffmann2022Body}. To address this, researchers have explored various sensing modalities, each with distinct advantages and trade-offs. The dominant approaches include creating large-area robotic skin to directly mimic biological touch and developing high-resolution vision-based sensors for detailed, localized contact analysis~\citep{Hardman2025Multimodal}. As a distinct alternative, vibro-acoustic sensing has emerged as a method for capturing the rich dynamics of interaction events, leveraging sound and vibration~\citep{Lee2025SonicBoom}. A further skin-free alternative estimates contact intrinsically from high-resolution joint-torque sensing: \citet{Iskandar2024Intrinsic} localize touch trajectories on a manipulator without any added surface instrumentation. However, this requires accurate torque sensing at every joint and is confined to the robot's own kinematic structure, whereas our microphones are inexpensive and decouple sensing density from the robot's actuation. This section provides an overview of the state-of-the-art approaches, highlighting their respective capabilities and limitations.

\subsection{Tactile Sensor Arrays and Robotic Skin}

Tactile sensor arrays and robotic skin technologies enable robots and prosthetic devices to sense and interpret touch, mimicking how human skin detects pressure, texture, and temperature. These systems are built from grids of sensitive elements embedded in flexible materials, enabling accurate, responsive detection of physical contact and environmental interactions. An example is Touchlab,\footnote{\url{https://www.touchlab.io/}, last accessed 2026-07-07.} which develops ultra-thin electronic skin. Their sensors equip robots with real-time touch sensing, enabling remote control with haptic feedback for healthcare and hazardous environments. \citet{Hardman2025Multimodal} developed another advanced tactile skin solution that uses a soft hydrogel embedded with over 860,000 conductive pathways to detect pressure, temperature, and damage, enabling highly sensitive, adaptive touch sensing for robots and prosthetics.
However, these methods face significant drawbacks. They are often expensive to produce and implement, and their delicate construction makes them fragile and susceptible to damage from routine physical interaction. Furthermore, the sensing capability is localized, meaning it is confined only to the specific areas where the robotic skin is applied, limiting the robot's overall environmental awareness~\citep{Chen2023How}.

\subsection{Vision-Based Tactile Sensing}
Vision-based tactile sensing takes a different approach by inferring contact information from camera observations of a deformable material. This approach can reconstruct detailed 3D contact geometry and force distribution from images. Examples of such sensors include GelSight~\citep{Yuan2017GelSighta}, which uses a camera to capture fine-grained texture and shapes of an illuminated elastomer. More compact designs such as DIGIT~\citep{Lambeta2020DIGIT} have helped standardize this approach and broaden its use in manipulation research. A key practical advantage of this family is simpler wiring, as a single camera replaces dense electrode arrays; we refer the reader to the recent taxonomy of \citet{Li2025VisionTactileReview} for a structured overview, and to biomimetic optical designs such as BioTacTip~\citep{Li2024BioTacTip} that recover 3D contact and force from raw tactile images. A primary advantage of this sensing family is detailed force reconstruction. Because the elastomer deforms according to elasticity theory, the observed deformation can be inverted to yield the continuous distribution of applied forces, rather than just a single resultant force vector. For instance, \citet{Ma2019DenseForce} recover a dense force distribution from tracked marker displacements using an inverse finite-element formulation, while \citet{Sferrazza2020Learning} reconstruct the full three-dimensional force distribution via a neural network trained in simulation and deployed on the physical sensor. These sensors provide rich data well-suited for grasping and in-hand manipulation, but they remain inherently local. Due to their bulk and reliance on internal optics, they are inherently difficult to scale for large surface areas and are not well-suited for whole-hand tactile sensing. Indeed, most dexterous hands with tactile sensing equip only the fingertips or finger pads (e.g., the Tactile SoftHand-A~\citep{Li2025SoftHand}), while the palm, dorsum, and forearm remain unsensed. Yet contacts during manipulation and unintended collisions frequently occur away from the fingertips, so whole-hand coverage is valuable for safe interaction and for detecting incidental contacts that fingertip-only sensing would miss. Vibro-acoustic sensing is attractive here precisely because a few microphones can cover an entire articulated surface without requiring individual sensors at every contact location. Therefore, the two sensor types represent a fundamental trade-off in spatial resolution: vision-based sensors resolve a small contact patch in high detail, yielding precise geometry and distributed force fields, whereas the vibro-acoustic approach presented here provides lower-resolution data about individual contacts, primarily detecting their location and trajectory.

\subsection{Vibro-Acoustic Sensing in Robotics}
Distinct from surface-based electronic or visual methods, vibro-acoustic sensing exploits structure-borne sound to detect physical interactions. Contact microphones convert mechanical vibrations propagating through the robot body into electrical signals, providing a lightweight, inexpensive, and highly scalable sensing modality. It can provide a complementary source of information to vision and force sensors, especially in occluded settings \citep{Toprak2018Evaluating}. In the following, we review relevant microphone-based work on (i) impulse response localization, which estimates contact locations from transient vibration patterns and (ii) trajectory estimation during contact-rich tasks, which tracks continuous motion from ongoing acoustic signals.

\subsubsection{Impulse Response Localization}
Instead of covering the robot with dense arrays, impulse localization estimates contact coordinates by decoding the structural vibrations captured by a sparse microphone array. For instance, SonicBoom~\citep{Lee2025SonicBoom} equips a robot end-effector with a distributed array of piezoelectric contact microphones to estimate the 3D locations of contact events. Using a data-driven approach that leverages relative acoustic features across microphones, SonicBoom achieves centimeter-level localization accuracy across varying surfaces and contact types, demonstrating the potential of passive acoustic sensing for high-resolution spatial perception even in occluded environments.

While passive methods capture naturally occurring contact vibrations, active vibro-acoustic methods introduce controlled vibrations into the robot or object to probe contact. \citet{Lu2023Active} integrated piezoelectric actuators with microphones in robotic grippers, enabling closed-loop estimation of contact state and object properties from reflected acoustic signals. Similarly, \citet{Wall2023Passive} embedded microphones into soft pneumatic actuators, enabling touch localization on deformable materials with complex geometries and in noisy environments. Such demonstrations highlight how acoustic sensing can augment contact perception without relying on dense tactile arrays or visual feedback.

Building on localization, the same vibration signals also reveal finer details, such as texture, slip, and material properties, making contact microphones highly complementary to other modalities. This richness enables biomimetic tactile designs, such as the fingerprint-inspired sensors from \citet{JuinaQuilachamin2023Fingerprint} that enhance spatial resolution and material identification. These capabilities naturally extend to tracking continuous interactions.

\subsubsection{Trajectory Tracking}
Although locating discrete impacts is well-studied, extending vibro-acoustic sensing to track the continuous motion of sliding or drawing interactions remains a significant challenge that has been addressed by fewer scientists.
\citet{Liu2024SonicSensea} proposed SonicSense, employing in-hand acoustic vibration sensing combined with deep neural networks to reconstruct detailed object motion information during manipulation. By analyzing vibration-induced acoustic patterns generated by sliding or interacting objects on gel surfaces, their system estimates motion direction, speed, and reconstructs 3D shapes using sound as the primary sensing modality.

In a related direction, \citet{Lu2020Multisensory} modeled and rendered tool-surface interaction sounds with wavelet-tree models segmented by contact velocity. This method, though primarily for perceptual rendering, demonstrates that velocity-dependent acoustic cues encode contact-motion information, underscoring the potential for trajectory inference from sound in manipulation tasks.

Extending this notion further, MilliSonic~\citep{Wang2019MilliSonic} achieves highly accurate acoustic motion tracking with sub-millimeter precision using airborne acoustic signals during free-space object motion. While this approach (i.e., airborne acoustic) lies outside the primary focus of this paper, it underscores the broader potential of acoustic modalities for precise trajectory estimation, even though it relies on external microphone arrays rather than contact-based vibration sensing.

These studies indicate that acoustic feedback contains rich spatiotemporal information to reconstruct continuous trajectories of hands or tools, opening new possibilities for audio-driven robotic control, high-resolution trajectory tracking, and dynamic, contact-rich manipulation.

The reviewed literature highlights that microphones can serve as lightweight and information-rich sensors for robotic interaction perception. However, the simultaneous application of this modality to both impulse response localization and continuous trajectory tracking, particularly while the robot itself is in motion, remains underexplored. Our work addresses this gap. In contrast to prior studies, we demonstrate that distributed contact microphones, coupled with a deep learning model, can robustly decode diverse contact behaviors. While our hardware setup shares the passive, multi-microphone design of \textit{SonicBoom}~\citep{Lee2025SonicBoom}, we significantly extend the scope to include dynamic trajectory localization, tracking external sliding contacts on the surface. Concretely, our novelty is threefold relative to the closest prior work: (i) we deploy sparse contact microphones on a complex, multi-degree-of-freedom anthropomorphic hand rather than on a rigid end-effector (SonicBoom~\citep{Lee2025SonicBoom}) or for in-hand object motion on a gripper (SonicSense~\citep{Liu2024SonicSensea}); (ii) we address both impulse-response localization and continuous trajectory tracking; and (iii) we analyze how material properties shape the vibrational signatures exploited by each task. Additionally, we provide a novel analysis of how material properties (e.g., stiffness vs.\ texture) distinctly influence the vibrational signatures used for localization. Furthermore, we validate our system's robustness to the robot's own actuator noise, a critical step toward enabling safe, whole-hand contact perception in real-world deployment.

\section{Method}
We propose a cost-effective yet accurate method that enables robots to perceive physical contact in a more natural manner. We demonstrate our approach on two real-world tasks: impulse response localization and a more complex case, trajectory tracking.

The following section describes the hardware setup of the robotic hand, followed by a detailed explanation of the two tasks.

\subsection{Hardware Setup}
\begin{figure}[htbp]
  \centering
  \includegraphics[height=65mm]{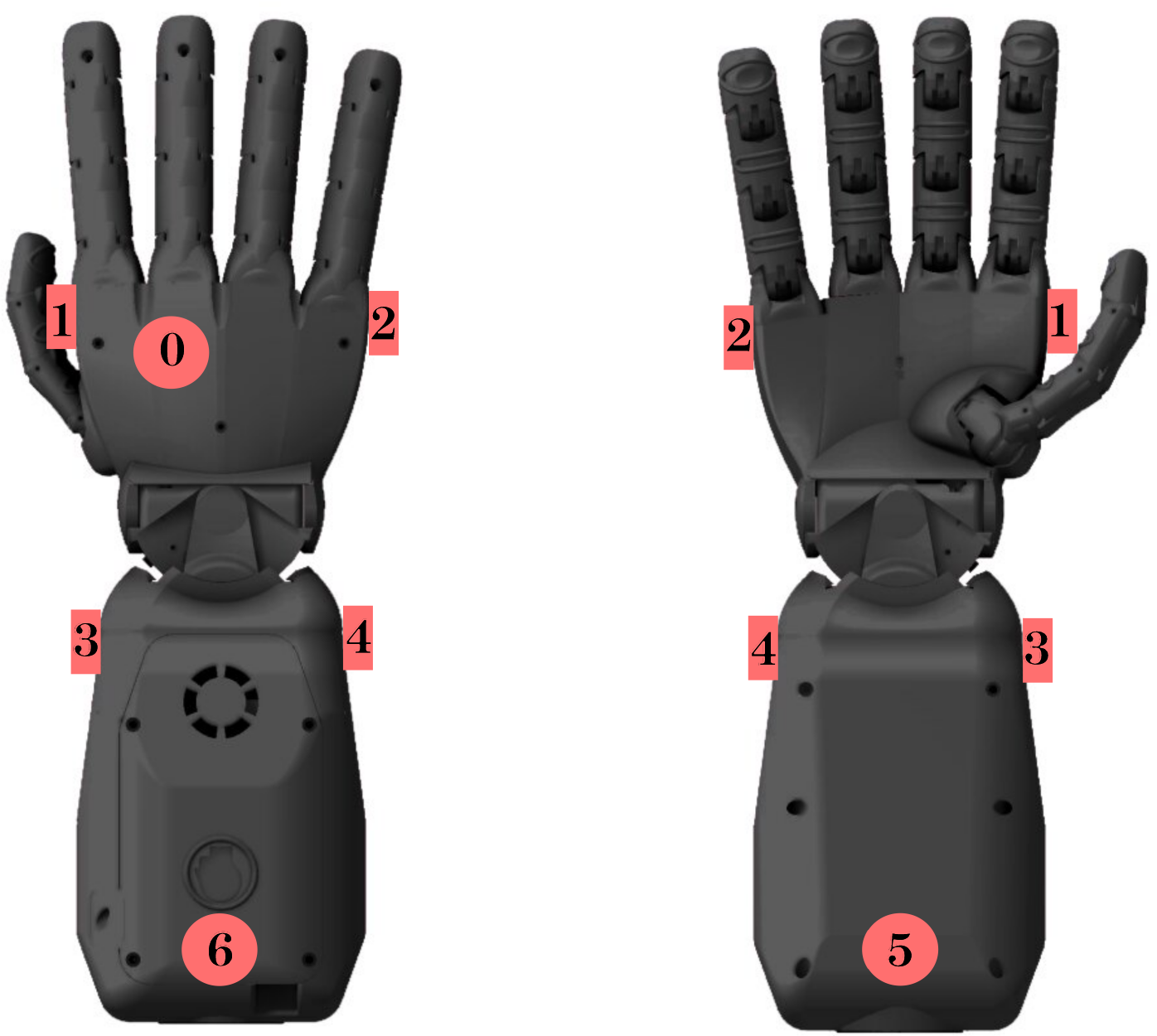}
  \caption{Schematic diagram illustrating the placement of microphones on the RH8D hand. The numbered markers denote the locations and specific IDs (0 through 6) of each microphone. Microphone 0 is mounted internally and faces the back side of the hand, while microphones 1 through 6 are mounted externally.}
  \label{fig:microphone-localization}
\end{figure}

For all our experiments, we used the Seed Robotics's RH8D hand~\citep{RH8D}. This hand has 19 degrees of freedom and is actuated by 8 motors. We equip the hand with seven Harley Benton CM-1000 contact microphones positioned to capture tactile vibrations (Fig.\ \ref{fig:microphone-localization}). Custom mounts were designed to fix the sensors in position; corresponding CAD models are provided on the project website. These microphones are capturing a range of $-500mV$ to $+500mV$.

\subsection{Impulse Response Localization Task}
\subsubsection{Task Overview}
For the impulse response localization task, we employed a UR5e robotic arm~\citep{UR5ea} equipped with a solenoid actuator, onto which we mounted four interchangeable cylindrical indenters made of distinct materials: soft plastic, hard plastic, wood, and metal, as shown in Figure~\ref{fig:experiment_setup}. The UR5e autonomously moves the solenoid to randomized positions facing the `Back', `Front', `Right', and `Left' sides of the robotic hand. Once in position, the solenoid is activated to deliver controlled poking interactions to the surface of the hand. A demonstration video of this data collection process is available on our website.

All contact positions are expressed in the base frame of the Seed RH8D hand. To obtain candidate contact points, we mount an Intel RealSense D435i depth camera and capture a depth map of each side of the hand. We deproject the depth map into a 3D point cloud and randomly sample target points from the hand surface. For each sampled point, the UR5e is commanded to approach along the surface normal and stop at a small standoff distance, after which the solenoid is energized to produce the impact. The ground-truth contact location is the commanded point: the UR5e tool-center-point pose is read from the controller and transformed from the robot base frame into the Seed hand base frame using the measured camera-to-robot and robot-to-hand rigid transforms. This yields an automatically labelled contact position for every sample without manual annotation.

The ground-truth localization is highly accurate, though some residual uncertainty remains. Two factors account for this: first, the depth camera operates at a scan range of $0.45$--$0.55\text{ m}$, where the manufacturer specifications for the D435i bound the depth error to a maximum of $2\,\%$ of the range; second, the indenter is driven onto the surface from a $12\text{ mm}$ standoff, meaning any misalignment between the contact direction and the local surface normal displaces the impact point laterally. Additionally, the UR5e pose repeatability of $\pm0.03\text{ mm}$ contributes to the total uncertainty, although its effect is negligible. Based on these compounding factors, we estimate the overall ground-truth uncertainty is minimal but non-negligible. Consequently, the evaluations presented in subsequent sections inherently include this ground-truth label noise, meaning any measured errors will represent an upper bound on the true performance.

\subsubsection{Dataset}
\label{sec:impulse_dataset}
The interactions generate mechanical vibrations and acoustic signals that propagate throughout the robotic hand's structure. Capturing these responses allows us to study the relationship between the resulting sensory feedback and the contact location. To build a robust dataset for training and evaluation, we collected approximately $65000$ unique samples.
Data were gathered from multiple sides and contact points around the robotic hand, both while the hand was idle and powered on and while it was off, to account for the background and internal noise introduced by the fan during normal operation.

Each interaction recording lasts $500\text{ ms}$, triggered $200\text{ ms}$ prior to contact. We initially captured raw signals at $50\text{ kHz}$, i.e.\ with a $25\text{ kHz}$ Nyquist bandwidth, to ensure full spectral coverage. To prepare this data for the neural network, we apply the following standardized preprocessing pipeline:

\begin{enumerate}
    \item \textbf{Downsampling:} The signals are resampled to a sampling rate of $20\text{ kHz}$, which preserves spectral content up to the corresponding Nyquist limit of $10\text{ kHz}$. As analyzed in Section~\ref{sec:param_optimization}, this rate was selected because sampling more finely, and thereby retaining a wider band, yielded no measurable gain in localization accuracy.
    \item \textbf{Windowing:} We trim the recording to a focused $200\text{ ms}$ window (from $125\text{ ms}$ to $325\text{ ms}$) to isolate the interaction event while discarding noise.
    \item \textbf{Feature Extraction:} The processed time-series data is converted into time--frequency features using a Short-Time Fourier Transform (STFT). We use a window size ($n_{\text{fft}}$) of 128, which was empirically determined to minimize localization error (see Section~\ref{sec:param_optimization}).
    \item \textbf{Noise Removal:} We utilize the discarded initial $100\text{ ms}$ of the pre-trigger phase to estimate and subtract the steady-state background noise.
\end{enumerate}

\subsection{Trajectory Tracking Task}
\subsubsection{Task Overview}
The trajectory tracking task is more complex, in which the UR5e robot arm draws different patterns on the surface of the forearm of Seed Robotics's RH8D hand~\citep{RH8D} using four different indenter materials. To automate the data collection procedure, we program the UR5e arm to draw a subset of real drawings from the open-source \textit{Quick Draw}~\citep{quickdraw} dataset, which spans 345 categories. We use the simplified version of the drawings. To minimize the robot's travel time between strokes, we optimized the drawing sequence by formulating the problem as a Generalized Traveling Salesperson Problem (GTSP)~\citep{Pop2024Comprehensive} and implemented this using the Google OR-Tools routing engine.\footnote{\url{https://developers.google.com/optimization}, last accessed 2026-07-07.} This approach allows each stroke to be drawn either forwards or in reverse and adjust its drawing order.

\subsubsection{Dataset}
For this task, we collect two datasets with different levels of complexity. The first dataset consists of 40,000 strokes per indenter (160,000 in total), with the Seed robotic hand powered on, idling in a fixed position, and its fan noise present. The second dataset contains 20,000 interactions per indenter (80,000 in total), during which the Seed robotic hand moves into random positions. Each stroke interaction varies in duration between 1s and 10s. This setup mimics real-world conditions, albeit exaggerated here to test the approach's limits, in which the robot performs actions while still needing to correctly sense its surroundings. The second dataset is therefore more challenging, as it requires the network to distinguish between hand motion and external touches.

Preprocessing follows the steps described in Section~\ref{sec:impulse_dataset}. We resample the signals to a $20\text{ kHz}$ sampling rate ($10\text{ kHz}$ Nyquist bandwidth). We then segment the continuous signals into fixed-length windows. For the target variable, we assign the average hand position $(x,y)$ recorded within each window. We then apply the same STFT parameters (128-window size) to generate consistent spectrogram features across both tasks. Finally, we remove the background noise using the first $100\text{ ms}$ of the recording.

Assigning this average position is a deliberate simplification. It is a reasonable target because the closely spaced stroke waypoints and the low blending radius limit the effective in-contact drawing speed to approximately $23\text{ mm/s}$ despite the commanded $0.25\text{ m/s}$, so the indenter moves only a few millimeters within a window; it nonetheless averages over intra-window motion and can smooth out local dynamics around sharp turns, fast strokes, or abrupt changes in contact. The window length controls this trade-off between temporal resolution and the number of vibrational cues per window, so we test several window sizes ($50\text{ ms}$, $100\text{ ms}$, $200\text{ ms}$, and $400\text{ ms}$); the resulting errors are reported later in Table~\ref{tab:window_ablation}. A fully sequence-based formulation (e.g., predicting a position per time step) is a promising direction we leave for future work.

\subsection{Networks Architecture}
We consider three network architectures: the Audio Spectrogram Transformer (AST)~\citep{Gong2021AST}, a ResNet-18 CNN~\citep{He2016Deep}, and a lightweight Temporal Convolutional Network (TCN)~\citep{Bai2018Empirical}. These are compared against each other and against a classical, non-neural $k$-nearest-neighbours baseline in Section~\ref{sec:baselines}. We employ the AST as the primary model due to its demonstrated effectiveness in learning from time--frequency representations of audio signals. AST leverages self-attention mechanisms to capture both local and global dependencies in spectrograms, making it particularly well-suited for tasks where temporal dynamics and subtle acoustic cues are critical. Its architecture allows for focusing on relevant patterns across the entire input, which is essential for accurately predicting target positions from audio interactions. Additionally, AST’s proven performance across a variety of audio classification and localization tasks ensures it generalizes well across our datasets, which vary in complexity.

 All three networks share the same input representation. We stack the synchronized spectrograms from the seven microphones along the channel dimension (creating a $7 \times T \times F$ tensor, where $T$ denotes the number of time frames and $F$ the number of frequency bins), allowing each model to learn inter-channel spatial features, such as phase and amplitude differences. Because the standard AST, ResNet-18, and TCN are designed for single- or three-channel inputs, we adapt the first layer of each, namely the patch-embedding layer in the AST and the input convolution in the ResNet-18 and TCN, to accept the 7-channel tensor; in the AST, these inter-channel cues are fused directly during tokenization.
All three networks are trained with an identical protocol using the mean squared error (MSE) loss; the exact layer configurations, optimizer, learning rate schedules, and batch size are specified in Appendix~\ref{app:architecture}.

Figure~\ref{fig:architecture} provides an overview of the full pipeline: the seven microphone signals are downsampled to $20$\,kHz, segmented into $200$\,ms windows, and converted into a stacked 7-channel dB spectrogram, which is decoded into a contact position by one of the three networks.

\begin{figure*}[t]
  \centering
  \includegraphics[width=\textwidth]{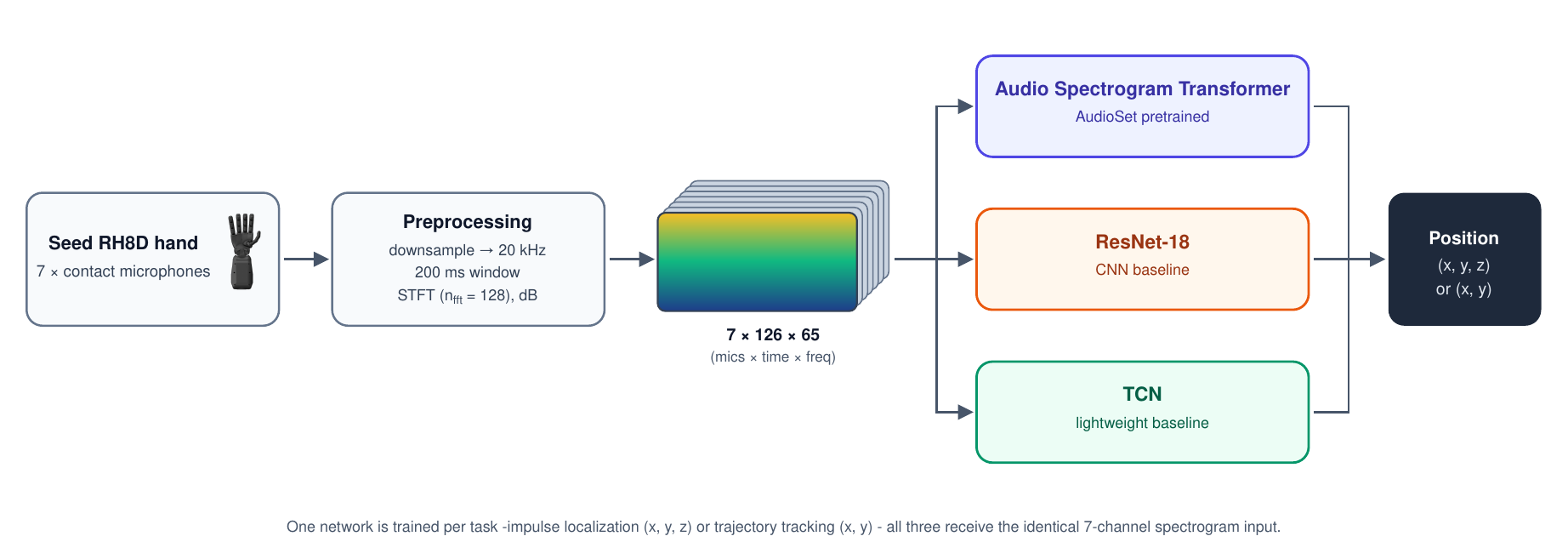}
  \caption{Overview of the sensing and decoding pipeline. Vibrations captured by the seven contact microphones on the RH8D hand are downsampled to $20$\,kHz, segmented into $200$\,ms windows, and transformed into a $7\times126\times65$ spectrogram tensor (microphones $\times$ time $\times$ frequency). The same input is decoded into a contact position by one of three neural architectures: the Audio Spectrogram Transformer (AST), a ResNet-18 CNN baseline, or a lightweight Temporal Convolutional Network (TCN). The output is $(x,y,z)$ for impulse-response localization or $(x,y)$ for trajectory tracking, with one network trained per task (the baselines are evaluated on impulse-response localization). Architectural and training details are given in Appendix~\ref{app:architecture}.}
  \label{fig:architecture}
\end{figure*}

\section{Experimental Results}
\subsection{Parameter Optimization \& Spectral Analysis} \label{sec:param_optimization}

\begin{figure}[htbp]
  \centering
  \includegraphics[width=\columnwidth]{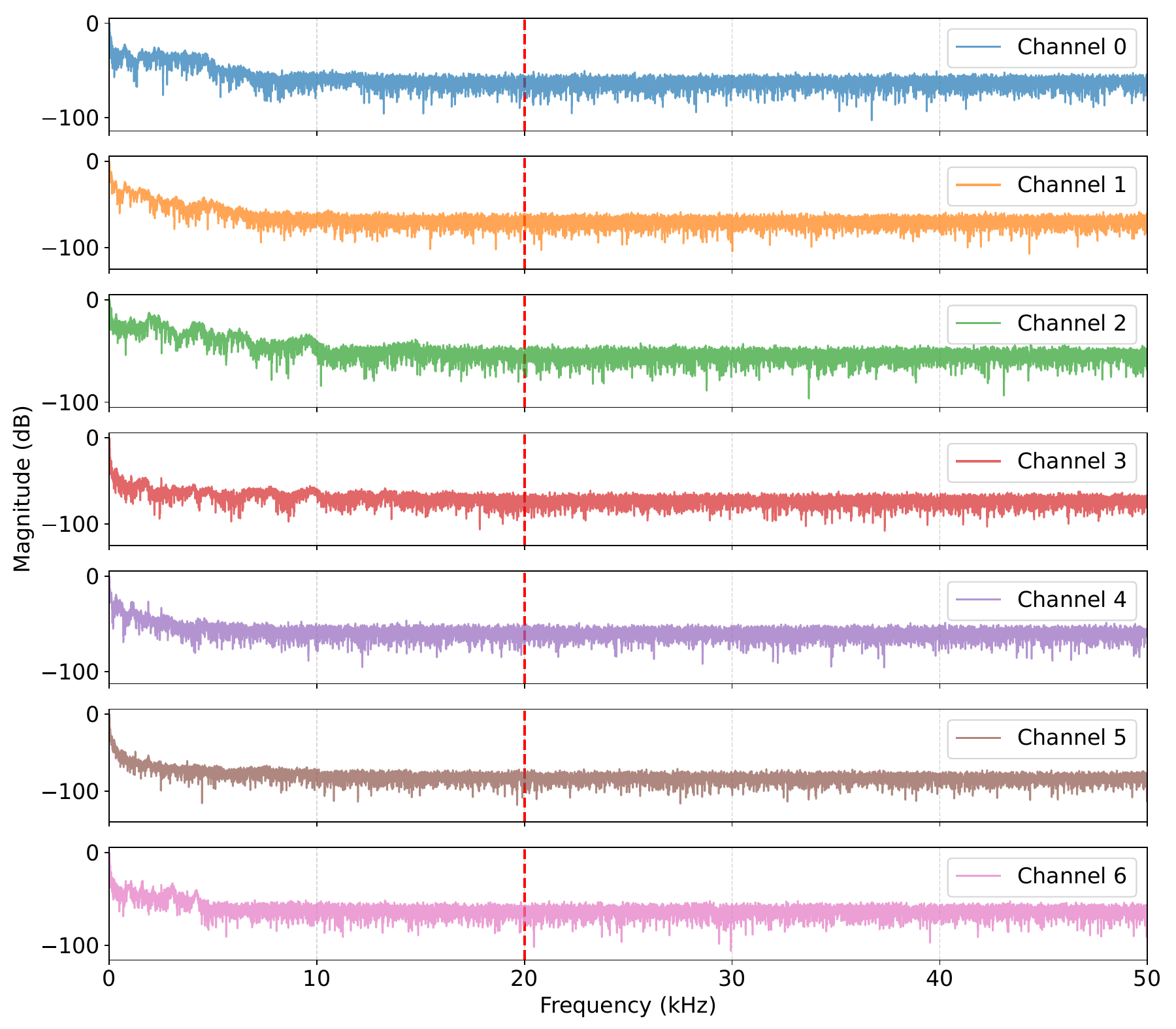}
  \caption{Magnitude spectra (in dB) of all seven channels. Each spectrum is normalized such that the maximum magnitude is 0 dB. Frequency is shown in kHz.}
  \label{fig:magnitude_frequencies}
\end{figure}

Before benchmarking the tasks' performance, we analyzed the spectral characteristics of the raw data to determine the optimal input frequency and STFT configuration. We first examined the frequency content of the raw $50\text{ kHz}$ signals. As illustrated in Figure~\ref{fig:magnitude_frequencies}, frequencies above $20\text{ kHz}$ exhibit negligible signal magnitude (below $-40\text{ dB}$). Beyond this point, the spectra of all channels remain flat at a uniformly low level, indicating that the informative content of the contact vibrations is concentrated below $20\text{ kHz}$. Notably, this holds for every microphone despite their different mounting positions on the hand, including the internally mounted one: no channel is limited by its placement, and all seven sensors capture the same informative frequency band. 

\begin{figure*}[tp]
  \centering
  \includegraphics[width=\textwidth]{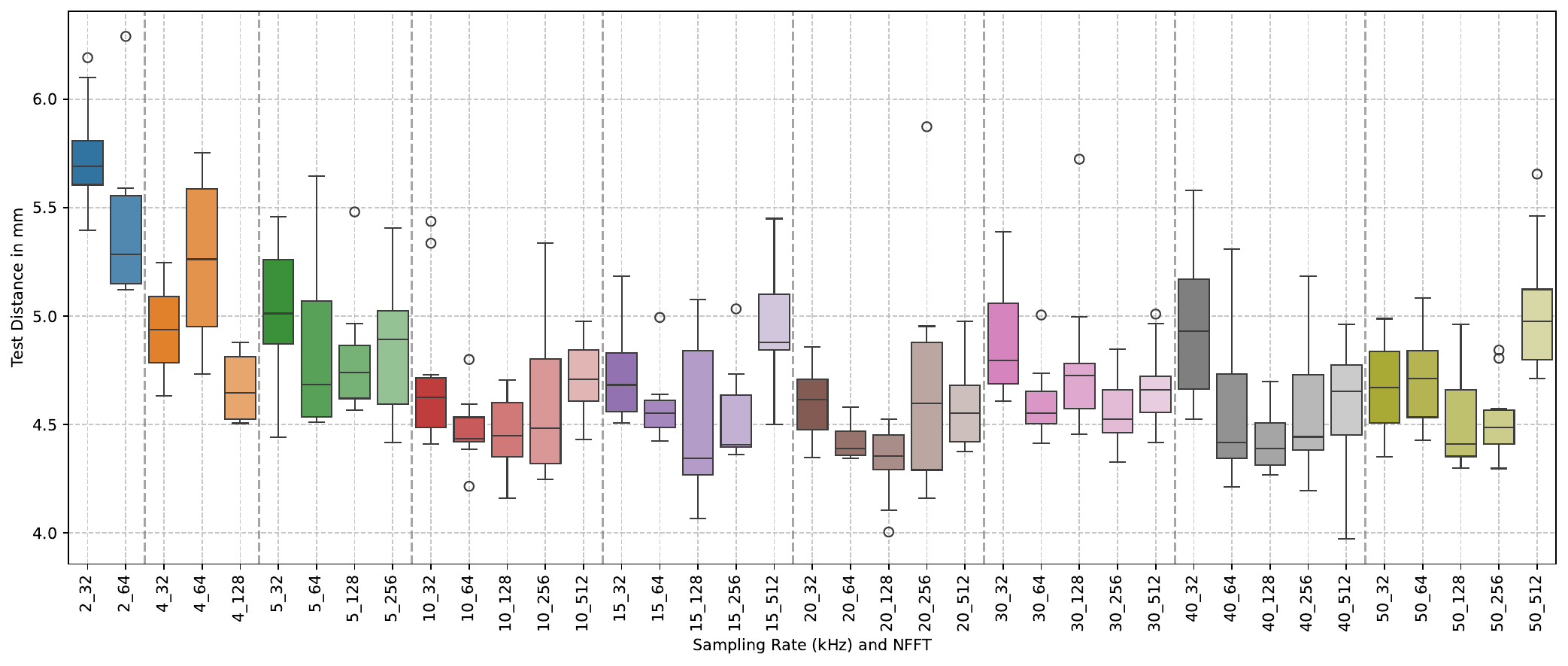}
  \caption{Test distance by sampling rate and window size ($n_{\text{fft}}$). Mean distances (mm) over 10 seeds per configuration, showing the influence of sampling rate and spectral resolution. The horizontal axis reports the sampling rate after downsampling, where the preserved bandwidth is half the stated value.}
  \label{fig:test_distance_freq_nfft}
\end{figure*}

To evaluate the impact of signal processing choices, we trained multiple neural networks (10 random seeds per configuration) while varying both the sampling rate after downsampling and the STFT window size ($n_{\text{fft}}$). Each sampling rate retains the full spectrum up to its Nyquist limit, that is, up to half the stated rate. For this parameter sweep, we utilized a fixed validation protocol: the `soft plastic' interactions were held out as the test set, and the networks were trained on the remaining materials. Figure~\ref{fig:test_distance_freq_nfft} reports the Euclidean distance (in mm) averaged across this held-out set over all 10 seeds.

The results indicate that sampling below $10\text{ kHz}$, which retains only components under $5\text{ kHz}$, degrades localization accuracy. However, sampling above $20\text{ kHz}$, and thereby admitting components above $10\text{ kHz}$, does not lead to any measurable performance improvements. Based on these findings, we selected a $20\text{ kHz}$ sampling rate, retaining a $10\text{ kHz}$ band, for our subsequent analyses. Alongside Figure~\ref{fig:magnitude_frequencies}, this confirms that the task-relevant content is restricted to frequencies under $10\text{ kHz}$: the vibrations do carry measurable energy between $10$ and $20\text{ kHz}$, but it adds no additional benefit to localization accuracy. This choice balances accuracy with computational efficiency by excluding frequencies that provide little or no useful information. Similarly, we chose a window size equal to 128, as it offered a favorable trade-off between time--frequency resolution and model performance in our preliminary evaluations, achieving an average Euclidean distance error of \textit{4.33} mm.

We note that the $20\text{ kHz}$ sampling rate, and the $10\text{ kHz}$ band it retains, was determined for our specific hand, materials, and microphone placement. While the informative band is unlikely to transfer, it is consistent with the frequency ranges reported for biomimetic contact sensing on different structures~\citep{JuinaQuilachamin2023Fingerprint}, suggesting the order of magnitude is broadly representative of structure-borne contact vibrations. For different robot geometries, materials, or sensor placements, however, the optimal bandwidth may shift and should be re-validated.

\subsection{Baseline Comparison}\label{sec:baselines}
To address whether the AST is necessary, we compare it against two alternative neural architectures and a classical feature-based baseline. As a strong convolutional baseline we adopt a ResNet-18 adapted to the seven-channel spectrogram input, and as a lightweight baseline the TCN specified in Appendix~\ref{app:architecture}; both are trained with the identical preprocessing pipeline, the same 10 random seeds, and the same `leave-one-material-out' protocol as the AST. Note that the AST is initialized from pretrained weights while both neural baselines are trained from scratch. As a non-neural reference, we add a $k$-nearest-neighbours (k-NN) regressor that predicts the distance-weighted average position of the closest training samples in a handcrafted feature space of per-channel spectral and peak-amplitude statistics (detailed in Appendix~\ref{app:architecture}), with the neighbourhood size selected on the validation split. Table~\ref{tab:baselines} reports the localization error of all four models on every material split.

As shown in Table~\ref{tab:baselines}, the AST outperforms both neural baselines on all four material splits, lowering the mean Euclidean error to $4.90$\,mm on average, compared with $7.90$\,mm for the ResNet-18 (a $38\%$ relative reduction) and $10.81$\,mm for the TCN (a $55\%$ relative reduction). The AST is also more consistent, with a smaller seed-to-seed standard deviation on the difficult splits.

The advantage of the AST is strongly material-dependent. It is comparable to the ResNet-18 on the acoustically ``easy'' stiff materials (Metal: $3.46$ vs.\ $4.06$\,mm; Soft Plastic: $4.94$ vs.\ $5.52$\,mm), but the gap widens sharply on the harder textured materials, where the vibrational signal is more complex: on Hard Plastic the AST nearly halves the error ($5.39$ vs.\ $9.67$\,mm) and on Wood it more than halves it ($5.82$ vs.\ $12.37$\,mm). The lightweight TCN shows the same degradation in a more pronounced form, its error roughly doubling from the stiff materials (Metal: $5.61$\,mm; Soft Plastic: $7.31$\,mm) to the textured ones (Hard Plastic: $14.59$\,mm; Wood: $15.74$\,mm).

While the transformer's global self-attention is a key feature, the improved performance on textured materials results from the design of the AST configuration. The AST and the ResNet-18 baseline represent distinct choices, differing in initialization, architecture, and parameter count ($86.5$\,M versus $11.2$\,M, Table~\ref{tab:latency}). Rather than isolating the effect of self-attention alone, Table~\ref{tab:baselines} demonstrates the practical superiority of the complete AST model under our training protocol. The results confirm that this specific combination of architecture, capacity, and pre-training provides a distinct advantage for acoustically complex materials. Fully ablating these individual components requires models with matched capacities, which we identify as a valuable direction for future work.

The classical k-NN baseline trails all neural models by a wide margin, averaging $16.12$\,mm (about $3.3\times$ the AST error), and its material ordering is inverted relative to the neural models, best on wood ($11.88$\,mm) and worst on soft plastic ($21.16$\,mm). We verified that this result is not an artefact of the feature choice: a linear ridge regressor on the same features and a PCA eigen-feature variant ($512$ components on the full spectrograms) performed no better ($18$--$23$\,mm on average), leaving the gap to the learned models unchanged.

\begin{table*}[ht]
\centering
\caption{Architecture comparison on impulse-response localization under the leave-one-material-out protocol. For the neural models (AST, ResNet-18, TCN), values are mean (std) over 10 seeds; the classical k-NN baseline (distance-weighted regression on pooled spectral features, Appendix~\ref{app:architecture}) is deterministic and reported as a single value. MSE is the normalized regression loss and the Euclidean distance is in mm.}
\label{tab:baselines}
\small
\setlength{\tabcolsep}{3.8pt}
\begin{tabular}{lcccccccc}
\toprule
 & \multicolumn{2}{c}{\textbf{AST}} & \multicolumn{2}{c}{\textbf{ResNet-18}} & \multicolumn{2}{c}{\textbf{TCN}} & \multicolumn{2}{c}{\textbf{k-NN}} \\
\cmidrule(lr){2-3}\cmidrule(lr){4-5}\cmidrule(lr){6-7}\cmidrule(lr){8-9}
\textbf{Test Split} & \textbf{MSE $\downarrow$} & \textbf{Dist. $\downarrow$} & \textbf{MSE $\downarrow$} & \textbf{Dist. $\downarrow$} & \textbf{MSE $\downarrow$} & \textbf{Dist. $\downarrow$} & \textbf{MSE $\downarrow$} & \textbf{Dist. $\downarrow$} \\
\midrule
Metal & $0.007 (0.000)$ & $3.46 (0.15)$ & $0.010 (0.001)$ & $4.06 (0.20)$ & $0.012 (0.001)$ & $5.61 (0.22)$ & $0.108$ & $16.69$ \\
Soft Plastic & $0.012 (0.001)$ & $4.94 (0.42)$ & $0.014 (0.001)$ & $5.52 (0.12)$ & $0.017 (0.001)$ & $7.31 (0.36)$ & $0.123$ & $21.16$ \\
Hard Plastic & $0.020 (0.007)$ & $5.39 (0.79)$ & $0.051 (0.012)$ & $9.67 (1.53)$ & $0.099 (0.016)$ & $14.59 (2.00)$ & $0.179$ & $14.75$ \\
Wood & $0.022 (0.011)$ & $5.82 (1.36)$ & $0.074 (0.018)$ & $12.37 (2.71)$ & $0.108 (0.016)$ & $15.74 (1.79)$ & $0.096$ & $11.88$ \\
\midrule
\textbf{Average} & \textbf{0.015} & \textbf{4.90} & $0.037$ & $7.90$ & $0.059$ & $10.81$ & $0.127$ & $16.12$ \\
\bottomrule
\end{tabular}
\end{table*}

Beyond accuracy, the three architectures differ sharply in computational cost. Table~\ref{tab:latency} reports the parameter count and the per-sample inference latency of each model (batch size 1, $7\times126\times65$ input), measured on a single GPU and on CPU. The AST processes a $200$\,ms window in $4.4$\,ms on GPU and $38.1$\,ms on CPU, comfortably within the real-time budget in both cases. The lightweight TCN baseline is over two orders of magnitude smaller and runs in $2.1$\,ms even on CPU, making it attractive for resource-constrained deployment despite its lower accuracy.

These values characterize the network alone, not the sensing system, whose latency is set by the $200$\,ms input window. Because a window must be captured before it can be processed, the end-to-end latency from contact to estimate is window-limited: $200$\,ms of signal, plus resampling, STFT and noise removal, plus a few milliseconds of inference, which is some $2$\,\% of the budget on GPU. For the impulse-response localization considered here, estimates are moreover produced per contact event rather than at a fixed rate, since each impact triggers the acquisition of the window. The system is therefore real time in the sense that it keeps pace with the incoming contacts.

\begin{table*}[ht]
\centering
\caption{Model size and per-sample inference latency (mean over 100 runs after warmup, batch size 1), measured on an NVIDIA RTX A6000 GPU and an AMD EPYC 7713 CPU.}
\label{tab:latency}
\begin{tabular}{lcccc}
\toprule
\textbf{Model} & \textbf{Params (M)} & \textbf{MACs (G)} & \textbf{GPU (ms)} & \textbf{CPU (ms)} \\
\midrule
AST & 86.5 & 5.42 & 4.4 & 38.1 \\
ResNet-18 & 11.2 & 0.39 & 2.0 & 7.6 \\
TCN (lightweight) & 0.2 & 0.03 & 1.3 & 2.1 \\
\bottomrule
\end{tabular}
\end{table*}

\subsection{Microphone Ablation}\label{sec:mic_ablation}
To assess how performance scales with the number of microphones, we vary the microphone count and report the AST error as the number of active microphones ranges from one to seven, supporting our claims of scalability and low-cost deployment. The microphone ablation is conducted on the soft-plastic held-out split, the same protocol used for parameter tuning in Section~\ref{sec:param_optimization}. Table~\ref{tab:mic_ablation} reports these results.

\begin{table*}[ht]
\centering
\caption{Microphone-count ablation for the AST on impulse-response localization, on the soft-plastic held-out split (values are mean (std) over 10 seeds). MSE is the normalized regression loss; the Euclidean distance (mm) is reported overall and split by contact height into `Forearm' ($z<150$\,mm) and `Hand' ($z\geq150$\,mm). Microphone subsets are listed as sensor indices, as presented in Fig.~\ref{fig:microphone-localization}.}
\label{tab:mic_ablation}
\begin{tabular}{lccccc}
\toprule
 & & & \multicolumn{3}{c}{\textbf{Euclid. Dist. (mm) $\downarrow$}} \\
\cmidrule(lr){4-6}
\textbf{Microphones} & \textbf{\# Mics} & \textbf{MSE $\downarrow$} & \textbf{Overall} & \textbf{Forearm} & \textbf{Hand} \\
\midrule
full (0--6) & 7 & \textbf{0.012 (0.001)} & \textbf{\phantom{0}4.94 (0.42)} & \textbf{\phantom{0}5.20 (0.51)} & \textbf{\phantom{0}4.54 (0.32)} \\
01234 & 5 & 0.021 (0.005) & \phantom{0}6.73 (1.45) & \phantom{0}7.61 (1.77) & \phantom{0}5.39 (1.01) \\
3456 & 4 & 0.016 (0.003) & \phantom{0}6.12 (1.08) & \phantom{0}6.63 (1.36) & \phantom{0}5.35 (0.69) \\
1234 & 4 & 0.025 (0.003) & \phantom{0}6.95 (0.87) & \phantom{0}7.88 (1.05) & \phantom{0}5.51 (0.64) \\
056 & 3 & 0.017 (0.004) & \phantom{0}6.13 (1.20) & \phantom{0}6.23 (1.30) & \phantom{0}5.97 (1.06) \\
012 & 3 & 0.041 (0.005) & \phantom{0}9.67 (1.35) & 11.12 (1.37) & \phantom{0}7.45 (1.37) \\
12 & 2 & 0.051 (0.004) & \phantom{0}9.72 (0.86) & 11.56 (0.90) & \phantom{0}6.90 (0.86) \\
0 & 1 & 0.061 (0.008) & 12.35 (1.38) & 13.63 (1.20) & 10.38 (1.71) \\

\bottomrule
\end{tabular}
\end{table*}

Table~\ref{tab:mic_ablation} shows that localization accuracy improves steadily as microphones are added: from $12.35$\,mm with a single microphone, the error drops to $9.72$\,mm with two, reaches the $6$--$7$\,mm range with three to five well-placed sensors, and is best with the full seven-microphone array ($4.94$\,mm). Placement contributes as much as the count: the three-microphone subset `056', already matches the four- and five-microphone subsets, whereas subsets confined to the hand (`012', `12') remain above $9.6$\,mm overall and their forearm error exceeds $11$\,mm. The internally mounted microphone 0 is the least informative on its own, and adding it to other subsets yields only marginal gains (e.g., $9.72$ to $9.67$\,mm when added to microphones 1 and 2), consistent with its weaker coupling to the outer shell. Finally, contacts on the hand section are localized more accurately than forearm contacts in every subset ($4.54$ vs.\ $5.20$\,mm for the full array), even though no microphone is mounted on the fingers themselves: the stiff finger and hand links transmit sharp, well-preserved transients to the surrounding microphones, whereas the larger forearm shell produces more diffuse vibration patterns.

To visualize how the microphones divide the sensing task spatially, Appendix~\ref{app:mic_coverage} maps every recorded contact location to the microphone with the strongest peak response. The microphones partition the hand and forearm into contiguous dominance regions centered on their mounting positions, and together they cover the entire surface. This spatial complementarity is consistent with the graceful degradation observed in Table~\ref{tab:mic_ablation} and with the strong performance of subsets that preserve coverage of both the hand and the forearm.

\subsection{Data Efficiency}\label{sec:data_efficiency}
\begin{figure}[htbp]
  \centering
  \includegraphics[width=\columnwidth]{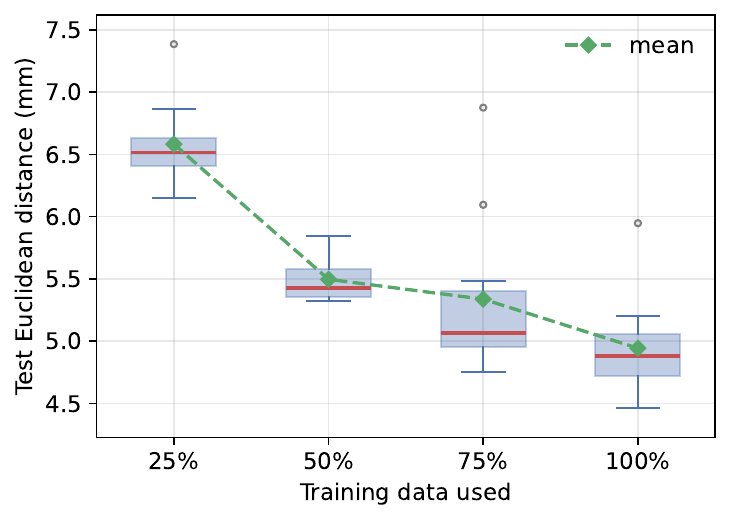}
  \caption{Data-efficiency analysis for impulse-response localization. For each training-data fraction, the boxplot shows the distribution of the test Euclidean error over 10 random seeds (median in red); the dashed line connects the per-fraction means.}
  \label{fig:data_efficiency}
\end{figure}

To assess how performance scales with the amount of training data, we retrain the AST for impulse-response localization on the soft-plastic held-out split (the same protocol as Section~\ref{sec:param_optimization}) using seeded random subsets comprising 25\%, 50\%, 75\%, and 100\% of the training data, keeping the validation and test sets fixed and averaging over 10 seeds per fraction. As reported in Table~\ref{tab:data_efficiency} and visualized in Fig.~\ref{fig:data_efficiency}, accuracy improves monotonically with more data: the mean Euclidean error decreases from 6.58\,mm at 25\% to 5.50, 5.34, and 4.94\,mm at 50\%, 75\%, and 100\%, respectively. Even a quarter of the training data already yields sub-7\,mm localization, and the full dataset, which we use in all other experiments, achieves the best accuracy. This sweep covers the impulse-response localization task; we did not repeat it for trajectory tracking, where every fraction would require retraining on the substantially larger stroke datasets, and we expect the qualitative trend to carry over. Since data collection is fully automated, enlarging either dataset costs robot time rather than manual annotation effort.

\begin{table}[ht]
\centering
\caption{Data-efficiency analysis for impulse-response localization on the soft-plastic held-out split: test error versus the fraction of training data used (values are mean (std) over 10 seeds; the validation and test sets are held fixed). MSE is the normalized regression loss; Euclidean distance is in mm.}
\label{tab:data_efficiency}
\begin{tabular}{lcc}
\toprule
\textbf{Training data} & \textbf{MSE $\downarrow$} & \textbf{Euclid. Dist. (mm) $\downarrow$} \\
\midrule
25\% & 0.016 (0.001) & 6.58 (0.34) \\
50\% & 0.013 (0.001) & 5.50 (0.20) \\
75\% & 0.013 (0.002) & 5.34 (0.66) \\
100\% & 0.012 (0.001) & 4.94 (0.42) \\
\bottomrule
\end{tabular}
\end{table}

\subsection{Impulse Response Localization Task}

To evaluate the model's ability to generalize across different surface properties, we extended our experimental analyses beyond the initial soft plastic evaluation scenario used for parameter tuning.
We adopted a ``leave-one-material-out'' strategy to test unseen material categories. Specifically, we conducted separate experiments in which soft plastic, hard plastic, wood, and metal were individually held out as the test set, while the network was trained on the remaining materials.
This approach rigorously tests the model's robustness to varying vibro-acoustic impedances and surface textures. Each configuration was trained using 10 different random seeds to ensure statistical reliability; the mean localization errors and standard deviations are summarized in Table~\ref{tab:results_localization}.

\begin{table}[ht]
\centering
\caption{Results on impulse response localization dataset: mean squared error (MSE) and Euclidean distance in mm across different test splits. The results are calculated over 10 different seeds. Lower MSE values indicate better performance. The values in parentheses represent the standard deviation.}
\label{tab:results_localization}
\begin{tabular}{lcc}
\toprule
\textbf{Test Split} & \textbf{MSE $\downarrow$} & \textbf{Euclid. Dist. $\downarrow$} \\
\midrule
\midrule
Metal & \textbf{0.007 (0.000)} & \textbf{3.46 (0.15)} \\
\midrule
Soft Plastic & 0.012 (0.001) & 4.94 (0.42) \\
\midrule
Hard Plastic & 0.020 (0.007) & 5.39 (0.79) \\
\midrule
Wood & 0.022 (0.011) & 5.82 (1.36) \\
\bottomrule
\end{tabular}
\end{table}

As presented in Table~\ref{tab:results_localization}, our findings reveal a strong correlation between the model's localization accuracy and the physical properties of the indenter material. This trend underscores that the model's performance is fundamentally tied to the distinct acoustic signature generated by each material's interaction with the surface. The lowest error (3.46 mm) was achieved with the metal indenter. Physically, the high stiffness and hardness of metals result in a sharp, high-energy impact that propagates efficiently, producing a clean and highly localized vibrational signal with minimal ambiguity.

Conversely, the wood indenter yielded the highest mean error (5.82 mm), followed by hard plastic (5.39 mm) and soft plastic (4.94 mm). This suggests that materials with lower density or acoustic impedance closer to that of the robotic shell itself may generate signals that are harder to distinguish from structural reverberations.

\begin{figure}[htbp]
  \centering
    \includegraphics[width=\columnwidth]{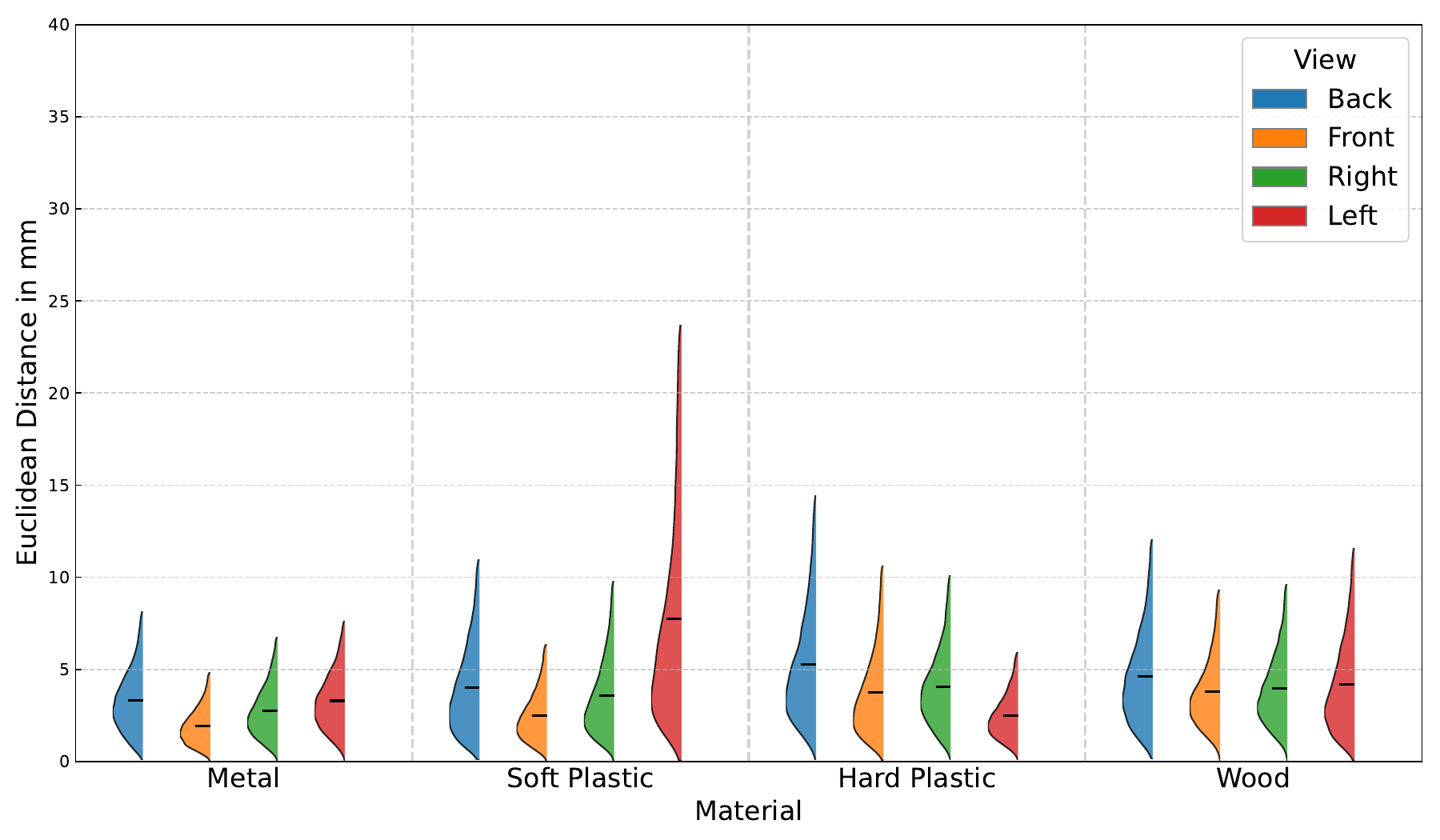} 
  \caption{Forearm localization error: half violin plots showing the distribution of prediction error (Euclidean Distance in mm) for four materials on the forearm section, broken down by view.}
  \label{fig:forearm_results}
\end{figure}

\begin{figure}[htbp]
  \centering
    \includegraphics[width=\columnwidth]{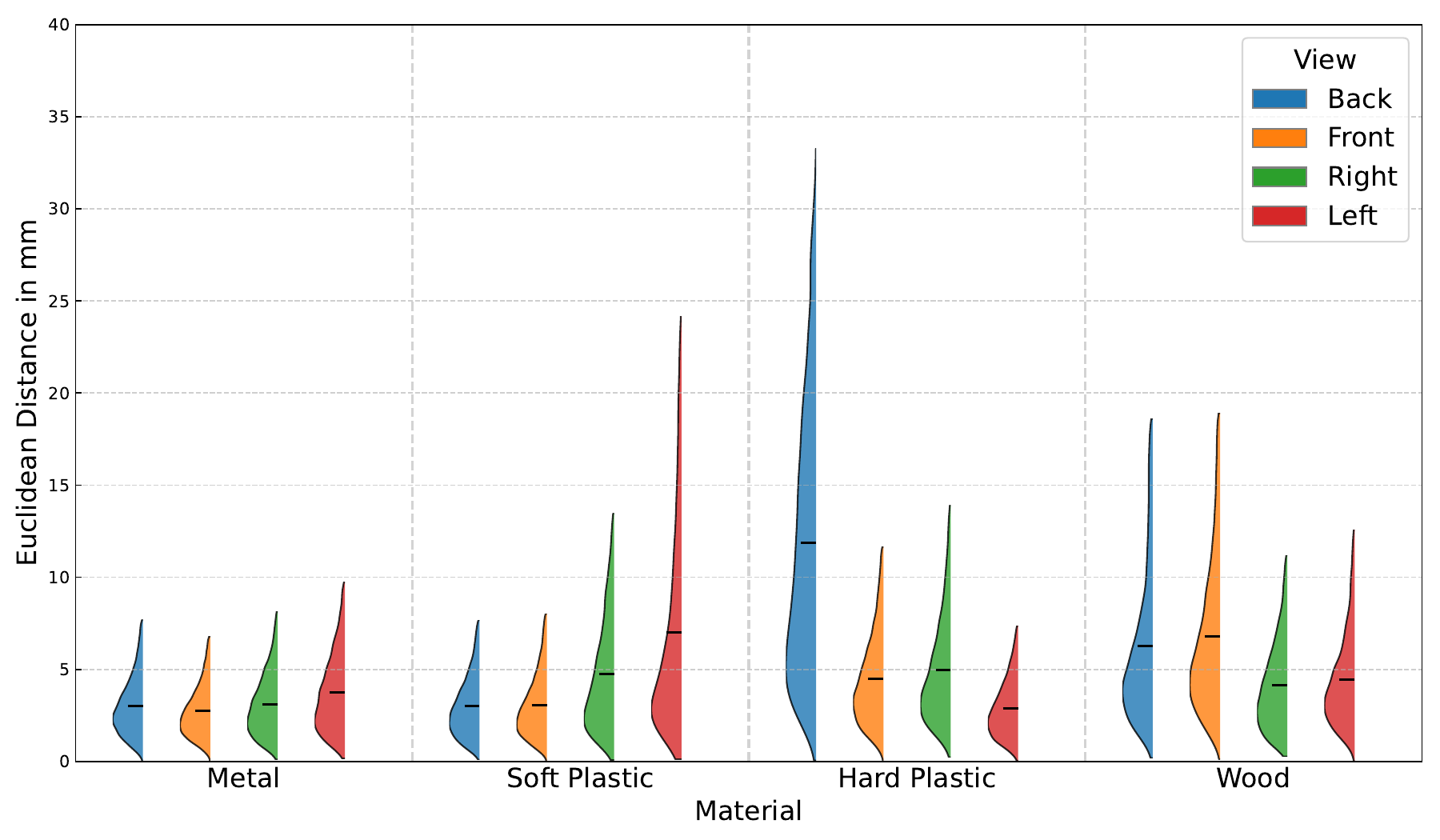} 
  \caption{Hand localization error: half violin plots showing the distribution of prediction error (Euclidean Distance in mm) for four materials on the hand section.}
  \label{fig:hand_results}
\end{figure}

To further investigate these results, we analyzed the per-view prediction error (Euclidean distance in mm) for the forearm (Fig.~\ref{fig:forearm_results}) and the hand (Fig.~\ref{fig:hand_results}). This visualization breaks down the error further by the `Back', `Front', `Right', and `Left' views.

The data indicates that the metal indenter provides the most spatially consistent acoustic signature for the impulse response localization task. As observed across all plots, metal achieves the lowest mean error and, notably, the lowest variance across all four views. The error distributions are tightly clustered near a low value (typically under 5 mm), suggesting that the high stiffness of the indenter minimizes deformation upon contact. This results in a sharp, high-energy signal transfer that remains consistent regardless of the contact location.

In comparison, the polymer and organic materials exhibit distinct view-dependent distributional characteristics:

\begin{itemize}
\item \textbf{Soft plastic:} The distribution for soft plastic highlights the influence of material compliance. We observe increased variance, specifically in the `Left' view, particularly in the Forearm plot. Physically, the softer material deforms upon impact, resulting in a damped interaction with less distinct transient features than with stiffer materials like metal. This lack of sharpness appears to make the signal more difficult to localize precisely, particularly on the left side of the robot, where structural geometry may further complicate signal propagation.

\item \textbf{Hard plastic:} While this material yields precise localization in most views, the analysis reveals a specific interaction dynamic in the `Back' view. As shown in the Hand (Fig.~\ref{fig:hand_results}) and Forearm (Fig.~\ref{fig:forearm_results}) configurations, the error distribution for this view extends noticeably (reaching 30-35 mm). This suggests that the acoustic impedance match between the hard plastic indenter and the robot's dorsal shell generates a signal propagation pattern that is more complex to spatially resolve than interactions on the ventral or lateral surfaces.

\item \textbf{Wood:} Finally, the wood indenter shows higher overall error but smaller distributional tails and less extreme outliers than soft plastic `Left' view and the hard plastic `Back' view. Compared to metal, it exhibits higher overall variance, particularly in the `Front' and `Back' views of the forearm. This characteristic variance aligns with the material's structural properties. Unlike metal and Plastic, which are manufactured to be homogeneous and isotropic, wood is a naturally heterogeneous material with variable density and stiffness across its geometry. These inherent irregularities introduce slight inconsistencies in the impact dynamics, resulting in a less deterministic acoustic signature than that of the uniform metal indenter. \end{itemize}

\subsection{Trajectory Tracking Task}\label{sec:trajectory_results}
For the trajectory tracking task, we evaluated our model using a similar `leave-one-material-out' approach. We report the AST only, with all seven microphones and the full training set, since this configuration performed best on the impulse response localization task (Sections~\ref{sec:baselines},~\ref{sec:mic_ablation} and~\ref{sec:data_efficiency}). We conducted four separate experiments, each time holding out one indenter material (soft plastic, hard plastic, wood, or metal) as the test set while training on the remaining three. The material is the held-out factor in these experiments: the network is trained on drawings of the three remaining materials, drawn from the Quick Draw categories~\citep{quickdraw}, and tested on the held-out material using a separately collected test set.

Furthermore, to assess the model's robustness against the hand's own motion, we evaluated all test splits under three conditions. The first was a stationary scenario, in which the hand was turned on and remained in a fixed position. The second was a dynamic scenario, with the hand moving to randomized poses between interactions. The final, mixed scenario, utilized a combined dataset from both conditions to evaluate generalization across different motion contexts.

We first assess how sensitive the tracking error is to the window length used to segment the continuous recordings. This study is run on the soft-plastic held-out split. In addition to holding out the material, we also split the Quick Draw categories between the three sets, using 276 of the 345 categories for training, 35 for validation, and 34 for testing, so that no drawing shape seen during training reappears at validation or test time; validation is further restricted to the three training materials. Table~\ref{tab:window_ablation} reports the test error in the `Fixed Position' and `Random Movement' scenarios when the dataset is regenerated with $50$\,ms, $100$\,ms, $200$\,ms, and $400$\,ms windows and the model retrained. The error decreases steadily as the window length increases in both scenarios. At the effective drawing speed of approximately $23$\,mm/s, a $50$\,ms or $100$\,ms window covers only $1$--$2$\,mm of travel, and the corresponding spectrograms contain 32 and 63 time frames versus 126 at $200$\,ms, so each sample carries substantially less vibrational evidence for the network to attend over; the sharper positional target of the short windows does not compensate for this weaker per-window signal.

Extending the window to $400$\,ms improves the `Fixed Position' error only marginally, within one standard deviation ($5.33$\,mm versus $5.68$\,mm), and reduces the `Random Movement' error from $13.79$\,mm to $11.21$\,mm, which we attribute to the longer integration time averaging out the transient ego-noise of the moving hand. This gain, however, comes at the cost of temporal fidelity: a $400$\,ms window halves the output rate from five to $2.5$ position estimates per second and averages the regression target over roughly $9$\,mm of travel, twice the $\approx 5$\,mm covered at $200$\,ms, thereby smoothing exactly the local dynamics that the windowing should preserve. We selected $200$\,ms to be able to cope with faster trajectory speeds.

\begin{table*}[ht]
\centering
\caption{Effect of the trajectory window length on tracking error: MSE and Euclidean distance in mm on the soft-plastic held-out split (values are mean (std) over 10 seeds). In addition to the held-out material, the Quick Draw categories are split between the three sets: 276 of the 345 categories for training, 35 for validation, and 34 for testing, with validation restricted to the training materials.}
\label{tab:window_ablation}
\begin{tabular}{llcc}
\toprule
\textbf{Window length} & \textbf{Scenario} & \textbf{MSE $\downarrow$} & \textbf{Euclid. Dist. $\downarrow$} \\
\midrule
\multirow{2}{*}{\phantom{0}$50\text{ ms}$}
  & Fixed Position & 0.133 (0.018) & \phantom{0}9.86 (0.68) \\
  & Random Movement & 0.267 (0.034) & 16.98 (0.99) \\
\midrule
\multirow{2}{*}{$100\text{ ms}$}
  & Fixed Position & 0.100 (0.015) & \phantom{0}8.54 (0.64) \\
  & Random Movement & 0.277 (0.066) & 16.11 (1.85) \\
\midrule
\multirow{2}{*}{$200\text{ ms}$}
  & Fixed Position & 0.046 (0.012) & \phantom{0}5.68 (0.69) \\
  & Random Movement & 0.204 (0.028) & 13.79 (0.97) \\
\midrule
\multirow{2}{*}{$400\text{ ms}$}
  & Fixed Position & \textbf{0.040 (0.013)} & \textbf{5.33 (0.82)} \\
  & Random Movement & \textbf{0.146 (0.013)} & \textbf{11.21 (0.45)} \\
\bottomrule
\end{tabular}
\end{table*}

We report the quantitative results in Table~\ref{tab:results_quickdraw}, including the mean squared error (MSE) loss and the Euclidean distance (in mm) calculated over the test splits for 10 different seeds. Note that these errors are lower than the corresponding $200$\,ms entries of the window study (Table~\ref{tab:window_ablation}): the window study also holds out the Quick Draw categories as described above, whereas the full evaluation trains on drawings from the full set of Quick Draw categories of the training materials and thus covers a broader range of stroke shapes during training.

\begin{table*}[ht]
\centering
\caption{Results on trajectory tracking dataset: mean squared error (MSE) and Euclidean distance in mm across different test splits and scenarios (10 seeds).}
\label{tab:results_quickdraw}
\begin{tabular}{llcc}
\toprule
\textbf{Test Split} & \textbf{Scenario} & \textbf{MSE $\downarrow$} & \textbf{Euclid. Dist. $\downarrow$} \\
\midrule
\midrule

\multirow{3}{*}{Metal}
  & Fixed Position & 0.015 (0.002) & 3.70 (0.19) \\
  & Random Movement & 0.130 (0.027) & 10.82 (1.20) \\
  & Fixed + Movement & 0.048 (0.004) & 5.64 (0.21) \\
\midrule
\multirow{3}{*}{Soft Plastic}
  & Fixed Position & 0.013 (0.001) & 3.47 (0.07) \\
  & Random Movement & 0.186 (0.018) & 12.95 (0.72) \\
  & Fixed + Movement & 0.085 (0.003) & 7.30 (0.17) \\
\midrule
\multirow{3}{*}{Hard Plastic}
  & Fixed Position & 0.008 (0.001) & 2.69 (0.12) \\
  & Random Movement & 0.177 (0.011) & 12.42 (0.39) \\
  & Fixed + Movement & 0.080 (0.007) & 6.75 (0.23) \\
\midrule
\multirow{3}{*}{Wood}
  & Fixed Position & \textbf{0.005 (0.001)} & \textbf{2.23 (0.08)} \\
  & Random Movement & \textbf{0.117 (0.015)} & \textbf{9.91 (0.72)} \\
  & Fixed + Movement & \textbf{0.042 (0.002)} & \textbf{4.83 (0.09)} \\
\bottomrule
\end{tabular}
\end{table*}

A critical finding in this experiment is the reversal of material performance rankings compared to the impulse response localization task, especially in the first test scenario with fixed position. While metal was the superior indenter for stationary poking, wood outperforms all other materials in this task, achieving the lowest error in every scenario (2.23 mm in `Fixed Position'), with soft/hard plastics performing in between. This shift highlights the fundamental physical difference between the two tasks. The impulse response localization task relies on impact dynamics, where metal's stiffness produces a sharp, high-bandwidth impulse. In contrast, the trajectory tracking task is driven by friction and surface interaction. The natural surface roughness and grain of the wood indenter generate a rich, continuous acoustic texture as it drags across the robot's surface, analogous to how biomimetic fingerprint ridges amplify structure-borne vibrations during sliding interactions~\citep{JuinaQuilachamin2023Fingerprint}, providing dense spectro-temporal features for tracking. Metal, being smoother, generates less distinct friction-induced vibrations when sliding, making the continuous path harder to reconstruct.

However, in the `Random Movement' and `Fixed + Movement' scenarios, hard and soft plastics show slightly higher errors than metal (e.g., Soft plastic: 12.95 mm vs.\ Metal: 10.82 mm in `Random Movement').
This may arise from excessive noise caused by minor, inconsistent hand movements across test set collections, where slight variations in individual hand poses can persist despite efforts to collect identical interactions.

Crucially, despite these challenges, the system maintains effective localization accuracy. Even in the extreme `Random Movement' condition, errors remain reasonably low (e.g., 9.91 mm for wood), validating that vibrational signal analysis is a robust sensing modality capable of providing robots with reliable physical awareness even during active operation.

\begin{figure}[htbp]
  \centering
    \includegraphics[width=0.8\columnwidth]{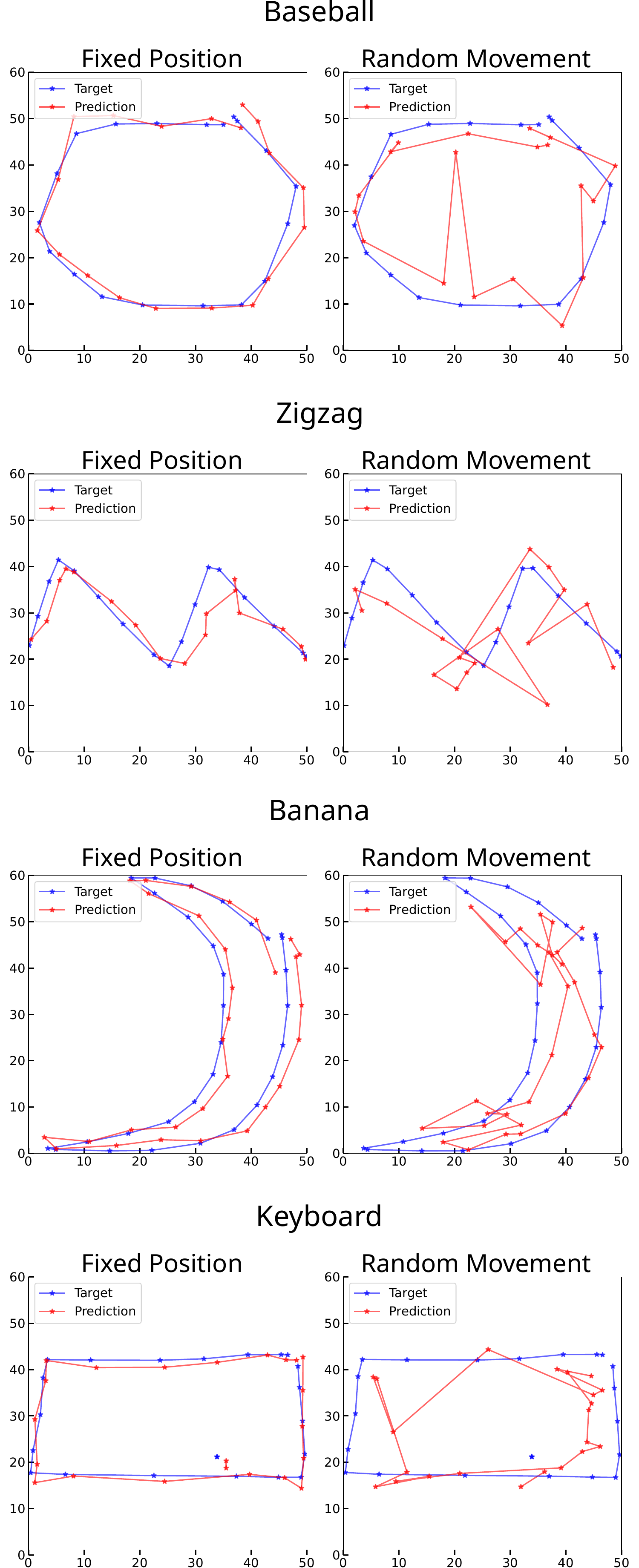} 
  \caption{Comparison of target (blue) vs.\ predicted (red) trajectories for fixed and random movement scenarios across four illustrative examples: Baseball, Zigzag, Banana, and Keyboard.}
  \label{fig:quickdraw examples}
\end{figure}

To contextualize these metrics, Figure~\ref{fig:quickdraw examples} visualizes trajectory reconstructions for four representative classes: Baseball, Zigzag, Banana, and Keyboard.

In the `Fixed Position' results, the model demonstrates high fidelity. The predicted path tightly matches the target, accurately capturing the continuous curvature of the Banana and Baseball as well as the sharp, distinct corners of the Keyboard. In the `Random Movement' scenario, the model retains the ability to track the underlying trajectory despite kinematic noise. Although the predictions exhibit increased variance and local deviations due to acoustic interference, the global topology of shapes like the Zigzag remains recognizable. The system recovers the overall structure of the path, identifying that the degradation is primarily manifested as jitter rather than a total loss of spatial coherence.

Based on these results, localization accuracy is governed by the interplay between the robot's kinematic state and the contact material properties. Across all material types, the `Fixed Position' experiment consistently yielded the highest precision. The `Random Movement' scenario, considerably more extreme than typical manipulation tasks, produced the highest error rates due to acoustic interference from internal motor actuation and structural vibrations. The `Fixed + Movement' scenario yielded intermediate performance, suggesting that the model can learn to partially generalize across varying noise conditions.

Finally, we assess whether the network can distinguish external contacts from the hand's self-generated vibrations, complementing the tracking results above with an explicit no-contact baseline. We extend the AST with a second, binary contact head attached to the shared backbone features and train both heads jointly: the contact head with a binary cross-entropy loss on every sample, and the $(x,y)$ regression head only on contact samples. The experiment follows the same leave-one-material-out protocol and data splits as the `Random Movement' scenario of Table~\ref{tab:results_quickdraw}: for each held-out material, the contact class is trained on the recordings of the remaining three materials and tested on the held-out material. Because the test trajectories stem from recordings separate from the training data, the contact training set additionally includes the test-session recordings of the three training materials, exposing the detector to acoustic variability. Because the test recordings are drawn from sessions strictly disjoint from the training data, the accuracies reported below inherently demonstrate successful cross-session generalization to both unseen materials and unseen recording sessions simultaneously. This robust cross-session operation is attributed to the multi-session training strategy. To illustrate the necessity of this approach, a control confirms that a network trained on a single session exceeds $99$\,\% validation accuracy within that session, yet misses between $71$\,\% and $93$\,\% of the contacts from a different session depending on the held-out material, with a false-positive rate below $0.5$\,\%. This finding demonstrates that a single-session model learns session-specific acoustics rather than a generalizable contact signature. By explicitly training across multiple sessions, our approach successfully regularizes the decision boundary and ensures robust contact detection. Importantly, this multi-recording inclusion introduces no data leakage: only recordings of the three training materials are added, while every recording of the held-out material, from all sessions, remains strictly excluded from training. The held-out material is therefore unseen at test time. The no-contact class consists of 1,882 recordings of the hand moving through randomized poses without any external touch, processed with the identical $200$\,ms window pipeline and divided into training, validation, and test recordings (21,054 training windows, approximately 4\% of the training data).

Across the four held-out materials (10 seeds each), the model reaches detection accuracies of $98.7$\,\% (metal), $99.3$\,\% (soft plastic), $99.6$\,\% (hard plastic), and $98.4$\,\% (wood), i.e.\ $99.0$\,\% on average, with an average false-positive rate of $0.2$\,\% (no-contact windows misclassified as contact) and an average false-negative rate of $0.8$\,\% (missed contacts of the unseen material). The tracking error of the jointly trained regression head ($6.37$, $7.83$, $6.31$, and $5.61$\,mm in the same material order) stays below the corresponding single-task results of Table~\ref{tab:results_quickdraw} ($10.82$, $12.95$, $12.42$, and $9.91$\,mm), a gain we attribute to the additional multi-recording training data rather than to multi-task learning itself. These tracking errors are therefore not directly comparable to Table~\ref{tab:results_quickdraw}, whose models are trained without these additional recordings. Overall, these results show that a single shared backbone can deliver contact detection and trajectory tracking at once without sacrificing accuracy on either task: the contact head does not degrade tracking, and the network neither hallucinates contacts during active self-motion nor misses them, remaining reliable even for materials never seen during training.

\section{Conclusion}

This paper demonstrates that high-accuracy touch localization on a robotic hand is achievable through vibrational signal analysis, offering a scalable, cost-effective alternative to complex tactile skin arrays. By leveraging an Audio Spectrogram Transformer (AST) to process vibrational signals from simple piezoelectric microphones, our method achieves robust performance across varied interaction modalities.

A key contribution of this work is the comprehensive analysis of how material properties and physical interaction modes influence vibro-acoustic sensing. Our results reveal a fundamental distinction between stationary and dynamic sensing: stiff materials (such as metal) generate the sharpest impulse responses for impulse-response localization, whereas textured materials (such as wood) produce the most distinct friction-based features for trajectory tracking. These findings highlight that acoustic sensing captures rich physical data beyond simple coordinates, intrinsically encoding the mechanical properties of the contact object.

We further validated the system's robustness in active scenarios. While the introduction of motion and internal motor noise inevitably degrades precision, the system maintains effective localization accuracy (with errors on the order of 10--13 mm even under extreme, arguably unrealistic conditions), indicating its potential for real-world tasks where robots must sense while grasping or manipulating objects. Moreover, a no-contact hand-motion baseline shows that the model can be trained to distinguish external contacts from internal motor and structural vibrations, with a low false-positive rate, addressing a key requirement for deployment during active motion. Our signal processing analysis confirms that a sampling rate of 20 kHz, which retains a $10\text{ kHz}$ band, is sufficient to capture these features, balancing computational efficiency and sensing fidelity.

Despite these findings, limitations remain. Our geometric analysis identified distinct acoustic interaction behaviors, including the impedance mismatch observed with hard plastic on the dorsal shell, and the signal damping caused by soft materials in complex structural areas. Additionally, although the system demonstrates resilience to self-generated motor noise, the localization error inevitably increases during motion.
We also delineate the scope of our generalization evaluation. Because contact points are sampled continuously and at random, the held-out test sets contain surface locations not present in training, so the reported errors already reflect generalization to unseen contact locations within the sampled region. However, it is not expected that the model extrapolates to regions of the hand never sampled, which is an inherent limitation of supervised regression. While our evaluation explores dynamic conditions, these findings primarily demonstrate robustness within the tested distributions rather than broader generalization. For instance, the `Random Movement' scenario establishes robustness to motion and ego-noise, but performance on entirely unseen poses remains untested. Similarly, the microphone ablation (Section~\ref{sec:mic_ablation}) evaluates sensor loss on a fixed set rather than physical sensor displacement. Generalization to unseen contact forces and speeds was also not studied here, as the solenoid actuation and the drawing speeds were held fixed. We identify these untested parameters, alongside controlled force and speed configurations, as important directions for future work.

Finally, our system targets contact perception rather than control: it provides the robot with real-time awareness of where and how it is being touched, but we do not close the perception-action loop by using these estimates to drive contact-aware behaviors. Demonstrating such closed-loop manipulation is a distinct problem that lies beyond the scope of this perception-focused study.
Future work will address these challenges by employing adaptive pre-filtering to decouple internal motor noise and fusing vibro-acoustic features with visual data to resolve geometric ambiguities. Furthermore, we aim to extend the system’s capabilities to include closed-loop manipulation and slip detection, thereby enhancing robustness during delicate and clutter-rich interaction tasks, and to validate generalization across poses, forces, and speeds in a real-time setting, including held-out hand poses and physically relocated microphones.

Finally, to facilitate reproducibility, we are making our model checkpoints, datasets, and experimental setups publicly available. By open-sourcing these resources, we aim to accelerate the development of affordable whole-hand contact perception, which is essential for advancing robotic grasping and manipulation capabilities.


\appendix

\section{Architecture and Training Details}\label{app:architecture}

This appendix specifies the three network architectures summarized in Figure~\ref{fig:architecture} and the training protocol shared by all of them. Every model receives the identical input: a $7\times126\times65$ tensor stacking the dB spectrograms of the seven microphones (126 time frames, 65 frequency bins per $200$\,ms window), and predicts for impulse-response localization ($x,y,z$) and for trajectory tracking ($x,y$). One network is trained per task; the ResNet-18, TCN, and classical k-NN baselines are evaluated on impulse-response localization, while the AST is used for both tasks.

\subsection{Audio Spectrogram Transformer}
The AST~\citep{Gong2021AST} splits the input into $16\times16$ patches with a stride of 10 in both the frequency and time dimensions (overlap of 6), yielding a $12\times5$ (time $\times$ frequency) grid of 60 overlapping patches. Each patch is linearly projected to a 768-dimensional embedding by a convolutional patch-embedding layer whose input is widened from 3 to 7 channels; the pretrained kernel weights are averaged over the original RGB channels and replicated across the seven microphone channels. Two learnable tokens ([CLS] and [DIST], inherited from DeiT~\citep{Touvron2021Training}) are used, and the pretrained positional embeddings are cropped and bilinearly interpolated to the 60-patch grid. The resulting sequence of 62 tokens is processed by 12 pre-norm transformer encoder blocks (embedding dimension 768, 12 attention heads, MLP hidden dimension 3072). The prediction is obtained by averaging the final [CLS] and [DIST] token embeddings and applying LayerNorm followed by a linear layer to $d$ outputs. The backbone ($\approx$87\,M parameters) is initialized from the AudioSet-pretrained AST checkpoint, which is itself initialized from an ImageNet-pretrained DeiT-Base/384.

\subsection{ResNet-18 Baseline}
The CNN baseline is a standard ResNet-18~\citep{He2016Deep} trained from scratch. The first convolution is replaced by a $7\times7$ convolution mapping the 7 input channels to 64 feature maps (stride 2, padding 3), followed by batch normalization, ReLU, and a $3\times3$ max-pooling with stride 2. The backbone then comprises the standard four stages of two BasicBlocks each, with 64, 128, 256, and 512 channels (stride 2 in stages 2--4). Global average pooling and a linear layer ($512 \rightarrow d$) produce the output ($\approx$11.2\,M parameters).

\subsection{Temporal Convolutional Network Baseline}
The lightweight TCN baseline~\citep{Bai2018Empirical} operates over the time axis: at each of the 126 time frames, the $7\times65$ microphone--frequency values are flattened into a 455-dimensional feature vector, giving a $455\times126$ sequence. The network stacks four dilated residual blocks with dilations 1, 2, 4, and 8; each block consists of two 1-D convolutions (kernel size 3, 64 channels, identical dilation), each followed by batch normalization, ReLU, and dropout (0.1), with a residual connection around the block (a $1\times1$ convolution matches the channel count in the first block). Temporal average pooling and a linear layer ($64 \rightarrow d$) produce the output ($\approx$0.2\,M parameters).

\subsection{Classical k-Nearest-Neighbours Baseline}
The classical baseline pools each $7\times65\times126$ dB spectrogram into a fixed handcrafted feature vector: the per-frequency-bin temporal mean and standard deviation ($2\times7\times65$), the per-frame spectral mean forming a temporal energy envelope ($7\times126$), and, for each microphone, the peak amplitude, its time of occurrence, and the overall standard deviation ($3\times7$), giving a $1813$-dimensional descriptor. Features are standardized with the training-set mean and variance, and a distance-weighted $k$-nearest-neighbours regressor predicts the $(x,y,z)$ position; the neighbourhood size $k$ (ranging from $5$ to $10$ across splits) is selected on the validation set. The regressor is deterministic given the training set, so no seed variance is reported. For completeness, we additionally evaluated a linear ridge regressor on the same features and a PCA variant ($512$ components fitted on the flattened spectrograms) for both regressors; none improved on the reported k-NN configuration (Table~\ref{tab:baselines}).

\subsection{Training Protocol}
The three neural models are trained with the identical protocol: batch size 128, AdamW without weight decay, and a cosine learning rate schedule with a linear warmup over the first 1\% of the training steps (peak learning rate $7\times10^{-4}$ for the AST, $1\times10^{-3}$ for the ResNet-18 baseline, and $5\times10^{-4}$ for the TCN). The MSE loss is computed on normalized coordinates: for localization, each axis is standardized with the training-set mean and standard deviation; for trajectory tracking, coordinates are min--max scaled to $[-1, 1]$ (the constant $z$ is dropped). Training uses gradient clipping (1.0), with early stopping on the validation loss. The checkpoint with the lowest validation loss is used for testing. Every configuration is repeated with 10 random seeds, and predictions are denormalized to millimetres before computing the Euclidean-distance metric.

\section{Spatial Coverage of the Microphone Array}\label{app:mic_coverage}

\begin{figure*}[htbp]
  \centering
  \includegraphics[width=.96\textwidth]{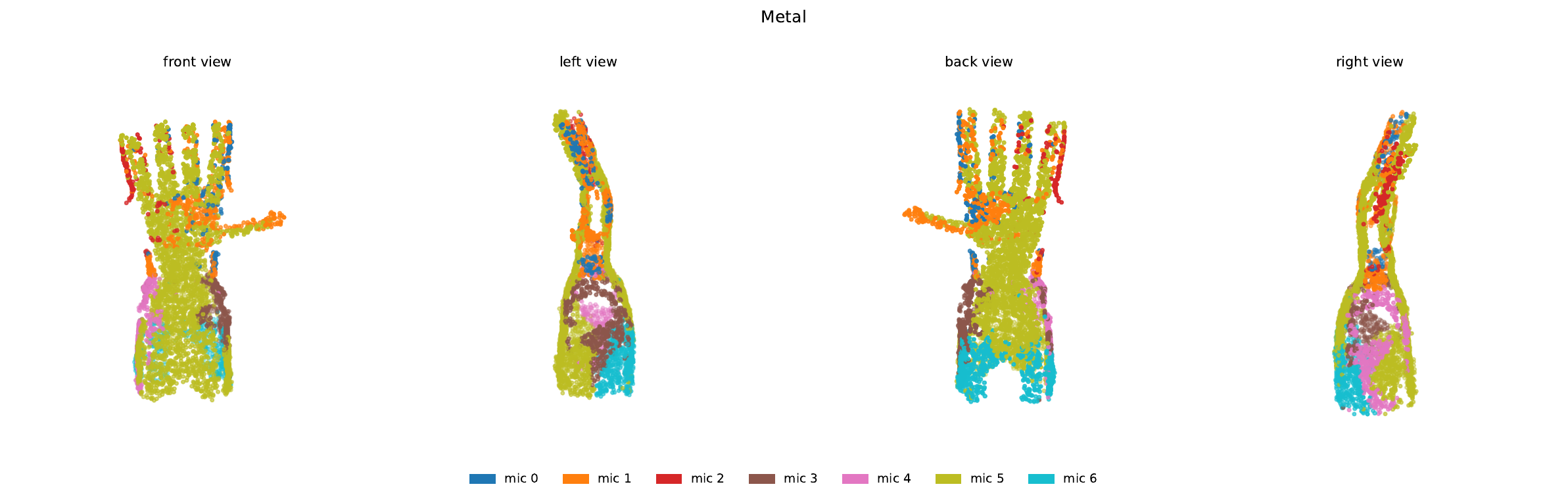}\\
  \includegraphics[width=.96\textwidth]{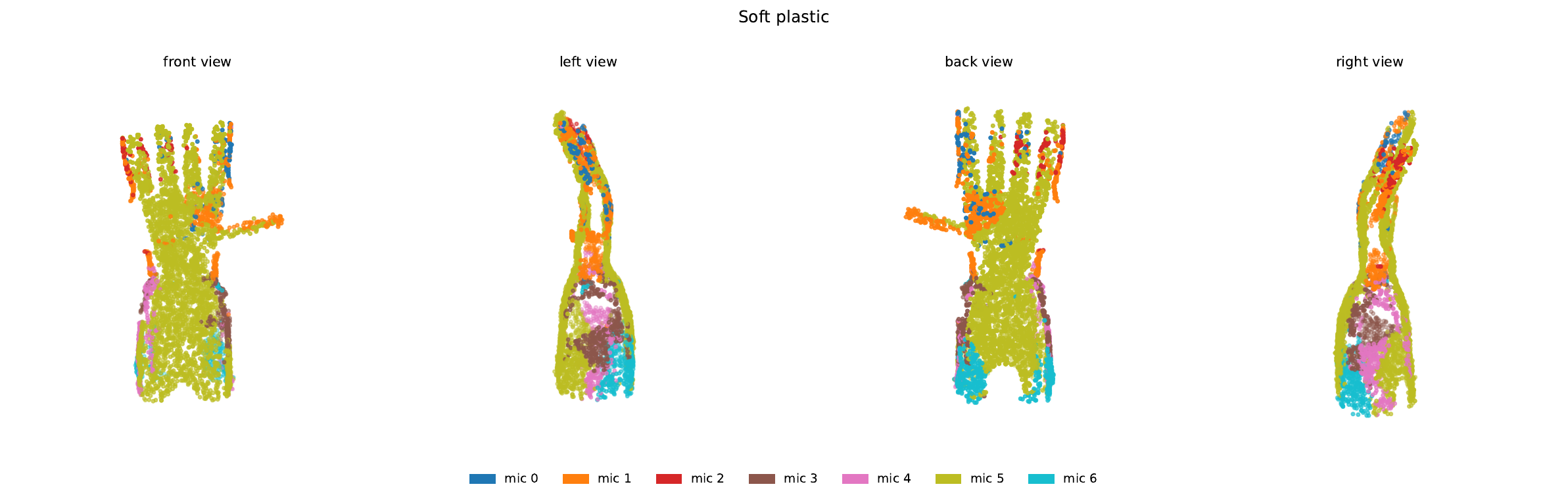}\\
  \includegraphics[width=.96\textwidth]{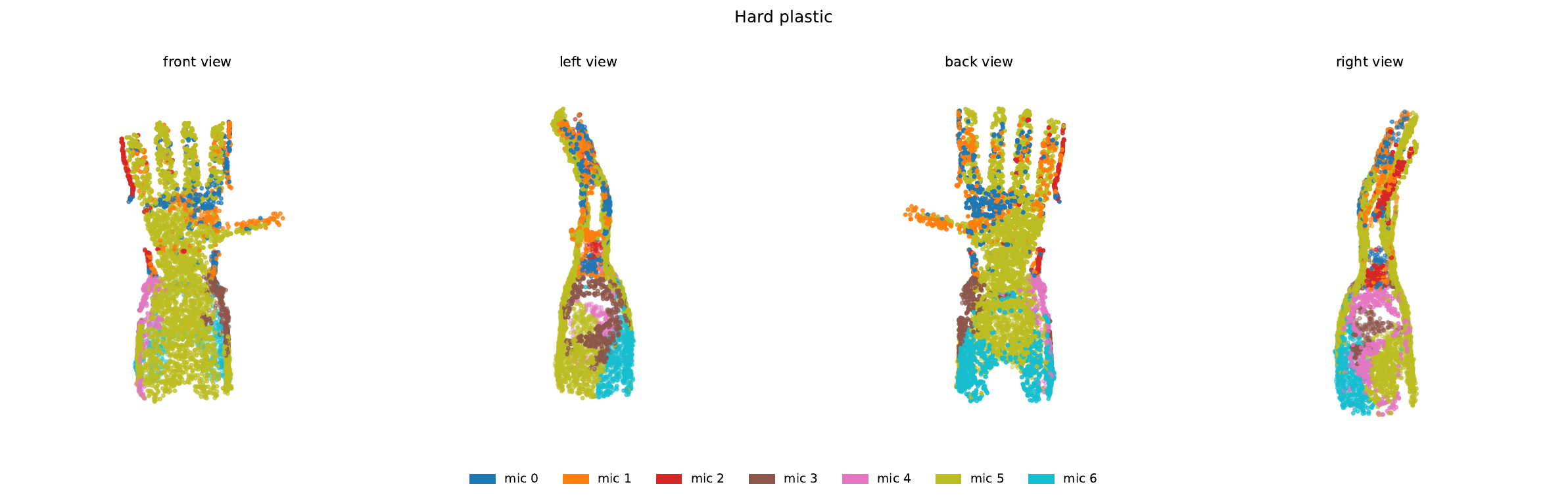}\\
  \includegraphics[width=.96\textwidth]{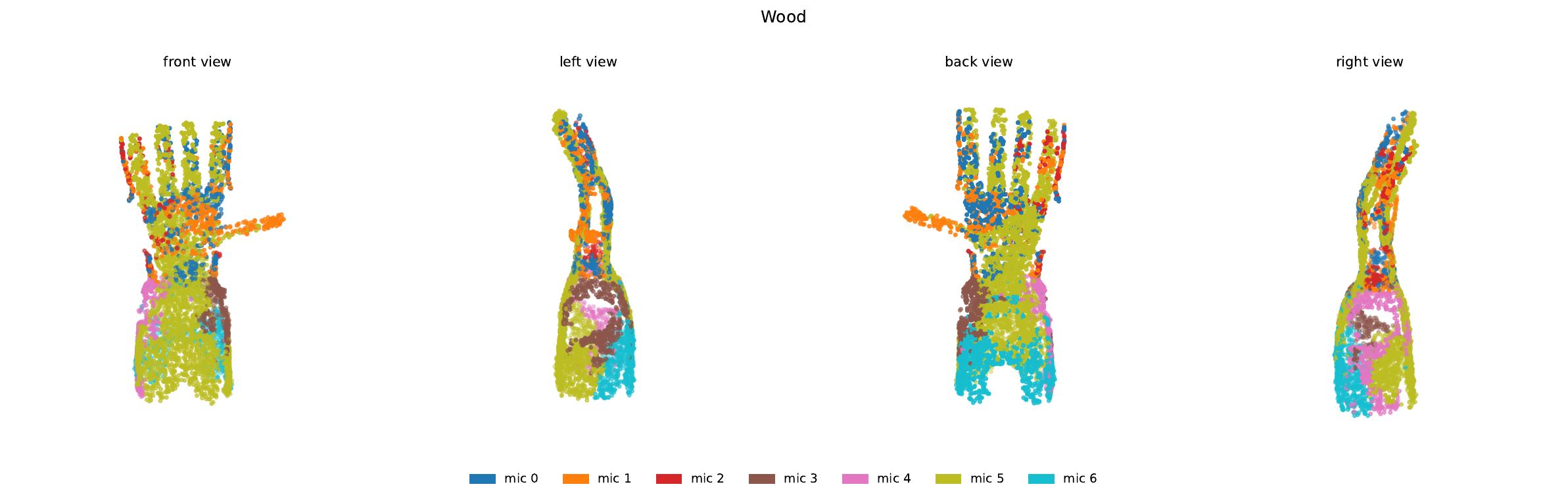}
  \caption{Spatial dominance maps of the microphone array for impulse-response localization. Each contact position (hand base frame) is colored by the microphone recording the largest peak amplitude, shown from four viewpoints (front, left, back, right) for each indenter material (microphone IDs as in Fig.~\ref{fig:microphone-localization}). The microphones partition the hand and forearm into contiguous dominance regions centered on their mounting positions, and every part of the surface is dominated by a nearby sensor.}
  \label{fig:mic_dominance}
\end{figure*}

To complement the microphone-count ablation of Section~\ref{sec:mic_ablation}, we analyze how the seven microphones divide the sensing task spatially. For every recorded poke of the impulse-response localization dataset, we compute the peak absolute amplitude of each microphone channel and assign the contact location to the microphone recording the largest peak amplitude. Figure~\ref{fig:mic_dominance} shows the resulting dominance maps. The externally mounted microphones partition the hand and forearm into contiguous dominance regions centered on their mounting positions (microphone IDs as in Fig.~\ref{fig:microphone-localization}). The forearm-base microphones 5 and 6 dominate the largest share of contact points: the base is a single merged shell that is structurally much simpler than the articulated hand and fingers, so its vibrations propagate over a wide area with little attenuation and are picked up strongly by these sensors. Contacts on the hand section are instead won by the hand-mounted sensors closest to them. The internally mounted microphone 0 is strongest only for a small number of scattered contact points, mostly on the fingertips; this is consistent with the ablation in Table~\ref{tab:mic_ablation}, where it is the least informative on its own and contributes only marginal gains to other subsets. Every region of the surface is thus covered by a nearby external sensor, confirming that the array provides complete spatial coverage and explaining why subsets that retain sensors on both the hand and the forearm degrade gracefully, whereas subsets confined to one section lose accuracy on the other.

\section*{Declarations}

\bmhead{Funding} This research was partially funded by the Bundesministerium Forschung, Technologie und Raumfahrt (BMFTR) under the Programme \href{https://www.bmbf.de/DE/Forschung/TransferInDiePraxis/DeutscheAgenturFuerTransferUndInnovation/Datipilot/datipilot_node.html}{DATIPilot} Innovationssprints project No.\ 03DPS1242A (\href{https://nicolas-navarro-guerrero.github.io/projects/vibrosense/}{Vibro-Sense}). 

\bmhead{Conflict of interest/Competing interests} The authors declare that they have no conflict of interest.
\bmhead{Ethics approval} This article does not contain any studies with human participants or animals performed by any of the authors.

\bmhead{Open Access} This article is licensed under a Creative Commons Attribution 4.0 International License, which permits use, sharing, adaptation, distribution, and reproduction in any medium or format, as long as you give appropriate credit to the original author(s) and the source, provide a link to the Creative Commons license, and indicate if changes were made. The images or other third-party material in this article are included in the article's Creative Commons license, unless indicated otherwise in a credit line to the material. If material is not included in the article's Creative Commons license and your intended use is not permitted by statutory regulation or exceeds the permitted use, you will need to obtain permission directly from the copyright holder. To view a copy of this license, visit \url{http://creativecommons.org/licenses/by/4.0/}.

\bmhead{Consent to participate} Not applicable
\bmhead{Consent for publication/Informed consent} Not applicable
\bmhead{Availability of data and materials}
Full dataset will be provided for the camera-ready version of this manuscript.
\bmhead{Code availability} Full code and checkpoints will be provided for the camera-ready version of this manuscript.

\bibliography{references.bib}


\end{document}